\documentclass{article}

\PassOptionsToPackage{numbers, compress}{natbib}

\usepackage[final]{neurips_2019}

\usepackage[utf8]{inputenc} 
\usepackage[T1]{fontenc}    
\usepackage{hyperref}       
\usepackage{url}            
\usepackage{booktabs}       
\usepackage{amsfonts}       
\usepackage{nicefrac}       
\usepackage{microtype}      
\usepackage{amsmath}
\usepackage{array}
\usepackage{graphicx}
\usepackage{amsthm}
\usepackage{subcaption}
\usepackage{arydshln}
\newcolumntype{L}[1]{>{\raggedright\let\newline\\\arraybackslash\hspace{0pt}}m{#1}}
\newtheorem{theorem}{Theorem}
\newtheorem{claim}{Claim}
\newtheorem{prop}{Proposition}

\newcommand*{\rom}[1]{\uppercase\expandafter{\romannumeral #1\relax}}
\newcommand{\inv}{^{\raisebox{.2ex}{$\scriptscriptstyle-1$}}}

\title{Self-attention with Functional Time Representation Learning}

\author{%
  Da Xu\thanks{The two first authors contribute equally to this work.} ,  Chuanwei Ruan\footnotemark[1] ,  Sushant Kumar ,  Evren Korpeoglu ,  Kannan Achan\\
  Walmart Labs\\
  California, CA 94086 \\
  \texttt{\{Da.Xu,Chuanwei.Ruan,EKorpeoglu,SKumar4,KAchan\}@walmartlabs.com} 
}

\begin{document}

\maketitle

\begin{abstract}
Sequential modelling with self-attention has achieved cutting edge performances 
in natural language processing. With advantages in model flexibility, computation complexity and interpretability, self-attention is gradually becoming a key component in event sequence models. However, like most other sequence models, self-attention does not account for the time span between events and thus captures sequential signals rather than temporal patterns. 
Without relying on recurrent network structures, self-attention recognizes event orderings via positional encoding. To bridge the gap between modelling time-independent and time-dependent event sequence, we introduce a functional feature map that embeds time span into high-dimensional spaces. By constructing the associated translation-invariant time kernel function, we reveal the functional forms of the feature map under classic functional function analysis results, namely Bochner's Theorem and Mercer's Theorem. We propose several models to learn the functional time representation and the interactions with event representation. These methods are evaluated on real-world datasets under various continuous-time event sequence prediction tasks. The experiments reveal that the proposed methods compare favorably to baseline models while also capturing useful time-event interactions. 
\end{abstract}

\section{Introduction}
\label{sec:introduction}
Attention mechanism, which assumes that the output of an event sequence is relevant to only part of the sequential input, is fast becoming an essential instrument for various machine learning tasks such as neural translation \citep{bahdanau2014neural}, image caption generation \cite{xu2015show} and speech recognition \cite{chorowski2015attention}. It works by capturing the importance weights of the sequential inputs successively and is often used as an add-on component to base models such as recurrent neural networks (RNNs) and convolutional neural networks (CNNs) \cite{chen2017sca}. 
Recently, a seq-to-seq model that relies only on an attention module called 'self-attention' achieved state-of-the-art performance in neural translation \cite{vaswani2017attention}. It detects attention weights from input event sequence and returns the sequence representation. Without relying on recurrent network structures, self-attention offers appealing computational advantage since sequence processing can be fully parallelized. Key to the original self-attention module is positional encoding, which maps discrete position index $\{1,\ldots, l\}$ to a vector in $\mathbb{R}^d$ and can be either fixed or jointly optimized as free parameters. Positional encoding allows self-attention to recognize ordering information. However, it also restricts the model to time-independent or discrete-time event sequence modelling where the difference in ordered positions can measure distance between event occurrences.  

In continuous-time event sequences, the time span between events often has significant implications on their relative importance for prediction. Since the events take place aperiodically, there are gaps between the sequential patterns and temporal patterns. For example, in user online behaviour analysis, the dwelling time often indicates the degree of interest on the web page while sequential information considers only the ordering of past browsing. Also, detecting interactions between temporal and event contexts is an increasingly important topic in user behavioural modelling \cite{li2017time}. In online shopping, transactions usually indicate long-term interests, while views are often short-termed. Therefore future recommendations should depend on both event contexts and the timestamp of event occurrences.

To effectively encode the event contexts and feed them to self-attention models, the discrete events are often embedded into a continuous vector space \cite{bengio2013representation}. After training, the inner product of their vector representations often reflect relationship such as similarity. In ordinary self-attention, the event embeddings are often added to positional encoding to form an event-position representation \cite{vaswani2017attention}. Therefore, it is natural and straightforward to think about replacing positional encoding with some functional mapping that embeds time into vector spaces. 

However, unlike positional encoding where representations are needed for only a finite number of indices, time span is a continuous variable. The challenges of embedding time are three folds. Firstly, a suitable functional form that takes time span as input needs to be identified. Secondly, the functional form must be properly parameterized and can be jointly optimized as part of the model. Finally, the embedded time representation should respect the function properties of time itself. To be specific, relative time difference plays far more critical roles than absolute timestamps, for both interpolation or extrapolation purposes in sequence modelling. Therefore, the relative positions of two time representations in the embedding space should be able to reflect their temporal difference. The contributions of our paper are concluded below:
\begin{itemize}
    \item We propose the translation-invariant \textsl{time kernel} which motivates several functional forms of \textsl{time feature mapping} justified from classic functional analysis theories, namely Bochner's Theorem \cite{loomis2013introduction} and Mercer's Theorem \cite{mercer1909xvi}. Compared with the other heuristic-driven time to vector methods, our proposals come with solid theoretical justifications and guarantees.
    \item We develop feasible \textsl{time embeddings} according to the \textsl{time feature mappings} such that they are properly parameterized and compatible with self-attention. We further discuss the interpretations of the proposed \textsl{time embeddings} and how to model their interactions with event representations under self-attention.
    \item We evaluate the proposed methods qualitatively and quantitatively and compare them with several baseline methods in various event prediction tasks with several datasets (two are public). We specifically compare with RNNs and self-attention with positional encoding to demonstrate the superiority of the proposed approach for continuous-time event sequence modelling. Several case studies are provided to show the time-event interactions captured by our model.
\end{itemize}

\section{Related Work}
\label{sec:related_work}
The original self-attention uses dot-product attention \cite{vaswani2017attention}, defined via:
\begin{equation}
    \text{Attn}(\mathbf{Q},\mathbf{K},\mathbf{V}) = \text{softmax}\Big(\frac{\mathbf{Q}\mathbf{K}^{\top}}{\sqrt{d}}\Big)\mathbf{V},
\end{equation}
where $\mathbf{Q}$ denotes the queries, $\mathbf{K}$ denotes the keys and $\mathbf{V}$ denotes the values (representations) of events in the sequence. Self-attention mechanism relies on the \emph{positional encoding} to recognize and capture sequential information, where the vector representation for each position, which is shared across all sequences, is added or concatenated to the corresponding event embeddings. The above $\mathbf{Q}$, $\mathbf{K}$ and $\mathbf{V}$ matrices are often linear (or identity) projections of the combined event-position representations. Attention patterns are detected through the inner products of query-key pairs, and propagate to the output as the weights for combining event values. Several variants of self-attention have been developed under different use cases including online recommendation \cite{kang2018self}, where sequence representations are often given by the attention-weighted sum of event embeddings.

To deal with continuous time input in RNNs, a time-LSTM model was proposed with modified gate structures \cite{zhu2017next}. Classic temporal point process also allows the usage of inter-event time interval as continuous random variable in modelling sequential observations \cite{zhao2015seismic}. Several methods are proposed to couple point process with RNNs to take account of temporal information \cite{xiao2017modeling, xiao2017joint, mei2017neural, du2016recurrent}. In these work, however, inter-event time intervals are directly appended to hidden event representations as inputs to RNNs. A recent work proposes a time-aware RNN with time encoding \cite{li2017time}.

The functional time embeddings proposed in our work have sound theoretical justifications and interpretations. Also, by replacing positional encoding with time embedding we inherit the advantages of self-attention such as computation efficiency and model interpretability. Although in this paper we do not discuss how to adapt the function time representation to other settings, the proposed approach can be viewed as a general time embedding technique.

\section{Preliminaries}
\label{sec:prelim}
Embedding time from an interval (suppose starting from origin) $T=[0, t_{\max}]$ to $\mathbb{R}^{d}$ is equivalent to finding a mapping $\Phi:T \to \mathbb{R}^{d}$. Time embeddings can be added or concatenated to event embedding $Z \in \mathbb{R}^{d_E}$, where $Z_i$ gives the vector representation of event $e_i$, $i=1,\ldots,V$ for a total of $V$ events. The intuition is that upon concatenation of the event and time representations, the dot product between two time-dependent events $(e_1, t_1)$ and $(e_2, t_2)$ becomes $\big[Z_1, \Phi(t_1)\big]^{'}\big[Z_2, \Phi(t_2)\big] = \big\langle Z_1, Z_2 \big\rangle + \big\langle \Phi(t_1), \Phi(t_2) \big\rangle$. Since $\langle Z_1, Z_2 \rangle$ represents relationship between events, we expect that $\big\langle \Phi(t_1), \Phi(t_2) \big\rangle$ captures temporal patterns, specially those related with the temporal difference $t_1 - t_2$ as we discussed before. This suggests formulating temporal patterns with a translation-invariant kernel $\mathcal{K}$ with $\Phi$ as the \textsl{feature map} associated with $\mathcal{K}$. 

Let the kernel be $\mathcal{K}:T \times T \to \mathbb{R}$ where $\mathcal{K}(t_1,t_2) := \langle \Phi(t_1), \Phi(t_2) \rangle$ and $\mathcal{K}(t_1, t_2) = \psi(t_1 - t_2), \forall t_1,t_2 \in T$ for some $\psi: [-t_{\max}, t_{\max}] \to \mathbb{R}$. Here the feature map $\Phi$ captures how kernel function embeds the original data into a higher dimensional space, so the idea of introducing the time kernel function is in accordance with our original goal.
Notice that the kernel function $\mathcal{K}$ is positive semidefinite (PSD) since we have expressed it with a Gram matrix. Without loss of generality we assume that $\Phi$ is continuous, which indicates that $\mathcal{K}$ is translation-invariant, PSD and also continuous.

So the task of learning temporal patterns is converted to a kernel learning problem with $\Phi$ as feature map. Also, the interactions between event embedding and time can now be recognized with some other mappings as $\big(Z, \Phi(t)\big) \mapsto f\big(Z, \Phi(t)\big)$, which we will discuss in Section \ref{sec:interaction}. By relating time embedding to kernel function learning, we hope to identify $\Phi$ with some functional forms which are compatible with current deep learning frameworks, such that computation via bask-propagation is still feasible. Classic functional analysis theories provides key insights for identifying candidate functional forms of $\Phi$. We first state Bochner's Theorem and Mercer's Theorem and briefly discuss their implications.

\begin{theorem}[Bochner's Theorem]
A continuous, translation-invariant kernel $\mathcal{K}(\mathbf{x},\mathbf{y})=\psi(\mathbf{x}-\mathbf{y})$ on $\mathbb{R}^d$ is positive definite if and only if there exists a non-negative measure on $\mathbb{R}$ such that $\psi$ is the Fourier transform of the measure.
\end{theorem}
The implication of Bochner's Theorem is that when scaled properly we can express $\mathcal{K}$ with:
\begin{equation}
\label{eqn:bochner}
    \mathcal{K}(t_1, t_2) = \psi(t_1, t_2) = \int_{\mathbb{R}}e^{i\omega(t_1 - t_2)}p(\omega)d\omega = E_{\omega}\big[\xi_{\omega}(t_1) \xi_{\omega}(t_2)^* \big],
\end{equation}
where $\xi_{\omega}(t) = e^{i\omega t}$. Since the kernel $\mathcal{K}$ and the probability measure $p(\omega)$ are real, we extract the real part of (\ref{eqn:bochner}) and obtain:
\begin{equation}
    \mathcal{K}(t_1,t_2) = E_{\omega}\big[\cos(\omega (t_1 - t_2))\big]=E_{\omega}\big[\cos(\omega t_1)\cos(\omega t_2) + \sin(\omega t_1)\sin(\omega t_2)\big].
\end{equation}

With this alternate expression of kernel function $\mathcal{K}$, the expectation term can be approximated by Monte Carlo integral \cite{rahimi2008random}. Suppose we have $d$ samples $\omega_1,\ldots,\omega_d$ drawn from $p(\omega)$, an estimate of our kernel $\mathcal{K}(t_1, t_2)$ can be constructed by $\frac{1}{d}\sum_{i=1}^d\cos(\omega_i t_1)\cos(\omega_i t_2) + \sin(\omega_i t_1)\sin(\omega_i t_2)$. As a consequence, Bochner's Theorem motivates the finite dimensional feature map to $\mathbb{R}^d$ via: 
\[
t \mapsto \Phi^{\mathcal{B}}_d(t) := \sqrt{\frac{1}{d}}\big[\cos(\omega_1 t),\sin(\omega_1 t),\ldots,\cos(\omega_d t), \sin(\omega_d t)\big],
\] 
such that $\mathcal{K}(t_1, t_2) \approx \lim_{d \to \infty}\big\langle \Phi^{\mathcal{B}}_d(t_1),\Phi^{\mathcal{B}}_d(t_2) \big\rangle$. 

So far we have obtained a specific functional form for $\Phi$, which is essentially a random projection onto the high-dimensional vector space of i.i.d random variables with density given by $p(\omega)$, where each coordinate is then transformed by trigonometric functions. However, it is not clear how to sample from the unknown distribution of $\omega$. Otherwise we would already have $\mathcal{K}$ according to the Fourier transformation in (\ref{eqn:bochner}). Mercer's Theorem, on the other hand, motivates a deterministic approach. 

\begin{theorem}[Mercer's Theorem]
Consider the function class $L^2(\mathcal{X}, \mathbb{P})$ where $\mathcal{X}$ is compact. Suppose that the kernel function $\mathcal{K}$ is continuous with positive semidefinite and satisfy the condition 
$\int_{\mathcal{X}\times \mathcal{X}} \mathcal{K}^2(x,z)d\mathbb{P}(x) d\mathbb{P}(y) \leq \infty$, then there exist a sequence of eigenfunctions $(\phi_i)_{i=1}^{\infty}$ that form an orthonormal basis of $L^2(\mathcal{X}, \mathbb{P})$, and an associated set of non-negative eigenvalues $(c_i)_{i=1}^{\infty}$ such that 
\begin{equation}
    \mathcal{K}(x,z) = \sum_{i=1}^{\infty} c_i \phi_i(x) \phi_i(z),
\end{equation}
where the convergence of the infinite series holds absolutely and uniformly.
\end{theorem}
Mercer's Theorem provides intuition on how to embed instances from our functional domain $T$ into the infinite sequence space $\ell^2(\mathbb{N})$. To be specific, the mapping can be defined via $t \mapsto \Phi^{\mathcal{M}}(t):= \big[\sqrt{c_1}\phi_1(t), \sqrt{c_2}\phi_2(t), \ldots\big]$, and Mercer's Theorem guarantees the convergence of $\big\langle \Phi^{\mathcal{M}}(t_1),\Phi^{\mathcal{M}}(t_2) \big\rangle \rightarrow \mathcal{K}(t_1,t_2)$. 

The two theorems have provided critical insight behind the functional forms of feature map $\Phi$. However, they are still not applicable.
For the feature map motivated by Bochner's Theorem, let alone the infeasibility of sampling from unknown $p(\omega)$, the use of Monte Carlo estimation brings other uncertainties, i,e how many samples are needed for a decent approximation. As for the feature map from Mercer's Theorem, first of all, it is infinite dimensional. Secondly, it does not possess specific functional forms without making additional assumptions. The solutions to the above challenges are discussed in the next two sections.

\section{Bochner Time Embedding}
\label{sec:bochner}
\begin{table}[]
    \centering
    \begin{tabular}{|L{4.2cm}|c|L{3.2cm}|L{3.2cm}|}
    \hline
        Feature maps specified by $\big[\phi_{2i}(t),\phi_{2i+1}(t)\big]$ &Origin & Parameters & Interpretations of $\omega$ \\ \hline
        $\Big[\cos\big(\omega_i(\mu)t\big), \sin\big(\omega_i(\mu)t\big)\Big]$ &Bochner's & $\mu$: location-scale parameters specified for the \textsl{reparametrization trick}. & $\omega_i(\mu)$: converts the $i^{th}$ sample (drawn from auxiliary distribution) to target distribution under location-scale parameter $\mu$.  \\ \hline 
        $\Big[\cos\big(g_{\theta}(\omega_i)t\big), \sin\big(g_{\theta}(\omega_i)t\big)\Big]$ & Bochner's & $\theta$: parameters for the inverse CDF $F\inv =g_{\theta}$. & $\omega_i$: the $i^{th}$ sample drawn from the auxiliary distribution. \\ \hline 
        $\big[\cos(\tilde{\omega_i}t), \sin(\tilde{\omega_i}t)\big]$ &Bochner's & $\{\tilde{\omega}\}_{i=1}^d$: transformed samples under non-parametric inverse CDF transformation. & $\tilde{\omega}_i$: the $i^{th}$ sample of the underlying distribution $p(\omega)$ in Bochner's Theorem. \\ \hline
        $\big[\sqrt{c_{2i,k}}\cos(\omega_j t),$ $\sqrt{c_{2i+1,k}}\sin(\omega_j t)\big]$ &Mercer's & $\{c_{i,k}\}_{i=1}^{2d}$: the Fourier coefficients of corresponding $\mathcal{K}_{\omega_j}$, for $j=1,\ldots,k$. & $\omega_j$: the frequency for kernel function  $\mathcal{K}_{\omega_j}$ (can be parameters). \\ \hline
        
    \end{tabular}
    \caption{The proposed functional forms of the feature map $\Phi=[\ldots, \phi_{2i}(t),\phi_{2i+1}(t), \dots]$ motivated from Bochner's and Mercer's Theorem, with explanations of free parameters and interpretation of $\omega$.}
    \label{tab:feature_maps}
\end{table}

A practical solution to effectively learn the feature map suggested by Bochner's Theorem is to use the '\textsl{reparameterization trick}' \cite{kingma2013auto}. Reparameterization trick provides ideas on sampling from distributions by using auxiliary variable $\epsilon$ which has known independent marginal distribution $p(\epsilon)$. 

For 'location-scale' family distribution such as Gaussian distribution, suppose $\omega \sim N(\mu, \sigma)$, then with the auxiliary random variable $\epsilon \sim N(0,1)$, $\omega$ can be reparametrized as $\mu + \sigma \epsilon$. Now samples of $\omega$ are transformed from samples of $\epsilon$, and the free distribution parameters $\mu$ and $\sigma$ can be optimized as part of the whole learning model. With Gaussian distribution, the feature map $\Phi^{\mathcal{B}}_d$ suggested by Bochner's Theorem can be effectively parameterized by $\mu$ and $\sigma$, which are also the inputs to the functions $\omega_i(\mu, \sigma)$ that transforms the $i^{th}$ sample from the auxiliary distribution to a sample of target distribution (Table \ref{tab:feature_maps}).
A potential concern here is that the 'location-scale' family may not be rich enough to capture the complexity of temporal patterns under Fourier transformation. Indeed, the Fourier transform of a Gaussian function in the form of $f(x)\equiv e^{-ax^2}$ is another Gaussian function. An alternate approach is to use inverse cumulative distribution function CDF transformation. 

Let $F\inv$ be the inverse CDF of some probability distribution (if exists), then for $\epsilon$ sampled from uniform distribution, we can always use $F\inv (\epsilon)$ to generate samples of the desired distribution. This suggests parameterizing the inverse CDF function as $F\inv \equiv g_{\theta}(.)$ with some functional approximators such as neural networks or \emph{flow-based} CDF estimation methods including \textsl{normalizing flow} \cite{rezende2015variational} and \textsl{RealNVP} \cite{dinh2016density} (see the Appendix for more discussions). As a matter of fact, if the samples are first drawn (following either transformation method) and held fixed during training, we can consider using non-parametric transformations. For $\{\omega_i\}_{i=1}^d$ sampled from auxiliary distribution, let $\tilde{\omega}_i = F\inv (\omega_i),i=1,2,\cdot,d$, for some non-parametric inverse CDF $F\inv$. Since $\omega_i$ are fixed, learning $F\inv$ amounts to directly optimize the transformed samples $\{\tilde{\omega}\}_{i=1}^d$ as free parameters.

In short, the Bochner's time feature maps can be realized with \textsl{reparametrization trick} or parametric/nonparametric inverse CDF transformation. We refer to them as Bochner time encoding. In Table \ref{tab:feature_maps}, we conclude the functional forms for Bochner time encoding and provides explanations of the free parameters as well as the meanings of $\omega$. A sketched visual illustration is provided in the left panel of Table \ref{tab:visualization}.  Finally, we provide the theoretical justification that with samples drawn from the corresponding distribution $p(w)$, the Monte Carlo approximation converges uniformly to the kernel function $\mathcal{K}$ with high probability. The upper bound stated in Claim 1 provides some guidelines for the number of samples needed to achieve a good approximation.

\begin{claim}
Let $p(\omega)$ be the corresponding probability measure stated in Bochner's Theorem for kernel function $\mathcal{K}$. Suppose the feature map $\Phi$ is constructed as described above using samples $\{\omega_i\}_{i=1}^d$, we have
\begin{equation}
    \text{Pr}\Big(\sup_{t_1,t_2 \in T}\big|\Phi^{\mathcal{B}}_d(t_1)^{'} \Phi^{\mathcal{B}}_d(t_2) - \mathcal{K}(t_1,t_2)\big| \geq \epsilon \Big) \leq 4\sigma_p \sqrt{\frac{t_{\max}}{\epsilon}}exp\Big(\frac{-d\epsilon^2}{32} \Big),
\end{equation}
where $\sigma_p^2$ is the second momentum with respect to $p(\omega)$.
\end{claim}
The proof is provided in supplement material.

Therefore, we can use $\Omega \big(\frac{1}{\epsilon^2} \log \frac{\sigma_p^2 t_{\max}}{\epsilon}\big)$ samples (at the order of hundreds if $\epsilon \approx 0.1$) from $p(\omega)$ to have $\sup_{t_1,t_2 \in T}\big|\Phi^{\mathcal{B}}_d(t_1)^{'} \Phi^{\mathcal{B}}_d(t_2) - \mathcal{K}(t_1,t_2)\big| < \epsilon$ with any probability.

\section{Mercer Time Embedding}
\label{sec:mercer}
Mercer's Theorem solves the challenge of embedding time span onto a sequence space, however, the functional form of $\Phi$ is unknown and the space is infinite-dimensional. To deal with the first problem, we need to make an assumption on the periodic properties of $\mathcal{K}$ to meet the condition in Proposition 1, which states a fairly straightforward formulation of the functional mapping $\Phi(.)$.

\begin{prop}
For kernel function $\mathcal{K}$ that is continuous, PSD and translation-invariant with $\mathcal{K}=\psi(t_1 - t_2)$, suppose $\psi$ is a even periodic function with frequency $\omega$, i.e $\psi(t)= \psi(-t)$ and $\psi\big(t+\frac{2k}{\omega}\big)=\psi(t)$ for all $t \in [-\frac{1}{\omega}, \frac{1}{\omega}]$ and integers $k \in \mathbb{Z}$, the eigenfunctions of $\mathcal{K}$ are given by the Fourier basis.
\end{prop}

The proof of Proposition 1 is provided in supplement material.

Notice that in our setting the kernel $\mathcal{K}$ is not necessarily periodic. Nonetheless we may assume that the temporal patterns can be detected from a finite set of periodic kernels $\mathcal{K}_{\omega}: T \times T \to \mathbb{R}, \omega \in \{\omega_1,\ldots,\omega_k \}$
, where each $\mathcal{K}_{\omega}$ is a continuous, translation-invariant and PSD kernel further endowed with some frequency $\omega$. In other words, we project the unknown kernel function $\mathcal{K}$ onto a set of periodic kernels who have the same properties as $\mathcal{K}$.

According to Proposition 1 we immediately see that for each periodic kernel $\mathcal{K}_{\omega_i}$ the eigenfunctions stated in Mercer's Theorem are given by: $\phi_{2j}(t)=1$, $\phi_{2j}(t) = \cos \big(\frac{j \pi t}{\omega_i}\big), \phi_{2j+1}(t) = \sin \big(\frac{j \pi t}{\omega_i} \big) \text{  for } j=1,2,\ldots$, with $c_i, i=1,2,\ldots$ giving the corresponding Fourier coefficients. Therefore we have the infinite dimensional Mercer's feature map for each $\mathcal{K}_{\omega}$:
\begin{equation*}
    t \mapsto \Phi^{\mathcal{M}}_{\omega}(t) = \Big[\sqrt{c_1},  \ldots,\sqrt{c_{2j}}\cos \big(\frac{j \pi t}{\omega}\big), \sqrt{c_{2j+1}} \sin \big(\frac{j \pi t}{\omega} \big), \ldots\Big],
\end{equation*}
where we omit the dependency of all $c_j$ on $\omega$ for notation simplicity.

One significant advantage of expressing $K_{\omega}$ by Fourier series is that they often have nice truncation properties, which allows us to use the truncated feature map without loosing too much information. It has been shown that under mild conditions the Fourier coefficients $c_j$ decays exponentially to zero \cite{widom1964asymptotic}, and classic approximation theory guarantees a uniform convergence bound for truncated Fourier series \cite{jackson1930theory} (see Appendix for discussions).
As a consequence, we propose to use the truncated feature map $\Phi^{\mathcal{M}}_{\omega,d}(t)$, and thus the complete Mercer's time embedding is given by:
\begin{equation}
    t \mapsto \Phi^{\mathcal{M}}_{d} = \big[\Phi^{\mathcal{M}}_{\omega_1,d}(t), \ldots, \Phi^{\mathcal{M}}_{\omega_k,d}(t)\big]^{\top}.
\end{equation}
Therefore Mercer's feature map embeds the periodic kernel function into the high-dimensional space spanned by truncated Fourier basis under certain frequency. As for the unknown Fourier coefficients $c_j$, it is obvious that learning the kernel functions $\mathcal{K}_{\omega}$ is in form equivalent to learning their corresponding coefficients. To avoid unnecessary complications, we treat $c_j$ as free parameters.


Last but not least, we point out that the set of frequencies $\{\omega_1, \ldots, \omega_k \}$ that specifies each periodic kernel function should be able to cover a broad range of bandwidths in order to capture various signals and achieve good approximation. They can be either fixed or jointly optimized as free parameters. In our experiments they lead to similar performances if properly initialized, such as using a geometrically sequence: $\omega_i = \omega_{\max} - (\omega_{\max} - \omega_{\min})^{i/k}, i=1,\ldots,k,$ to cover $[\omega_{\min}, \omega_{\max}]$ with a focus on high-frequency regions. The sketched visual illustration is provided in the right panel of Table \ref{tab:visualization}.

\begin{table}[htb]
    \centering
    \begin{tabular}{|c | c|}
    \hline
        Bochner time embedding & Mercer time embedding  \\ \hline 
        \includegraphics[scale=0.2]{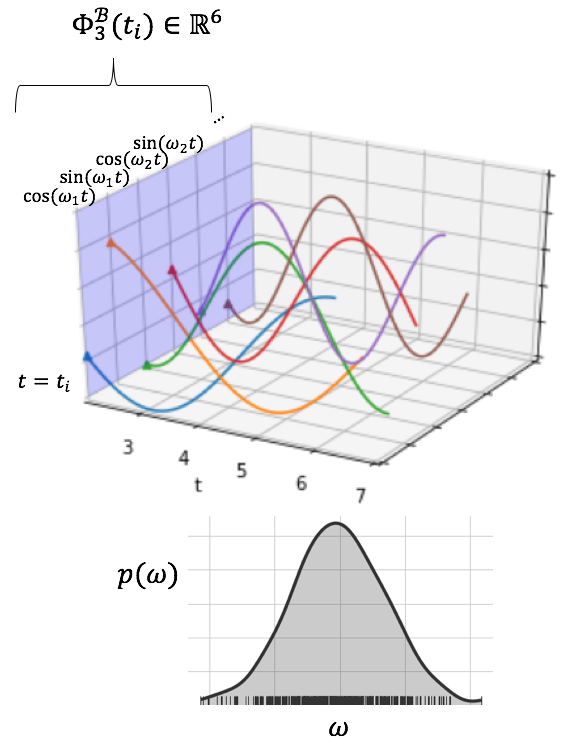} & \includegraphics[scale=0.2]{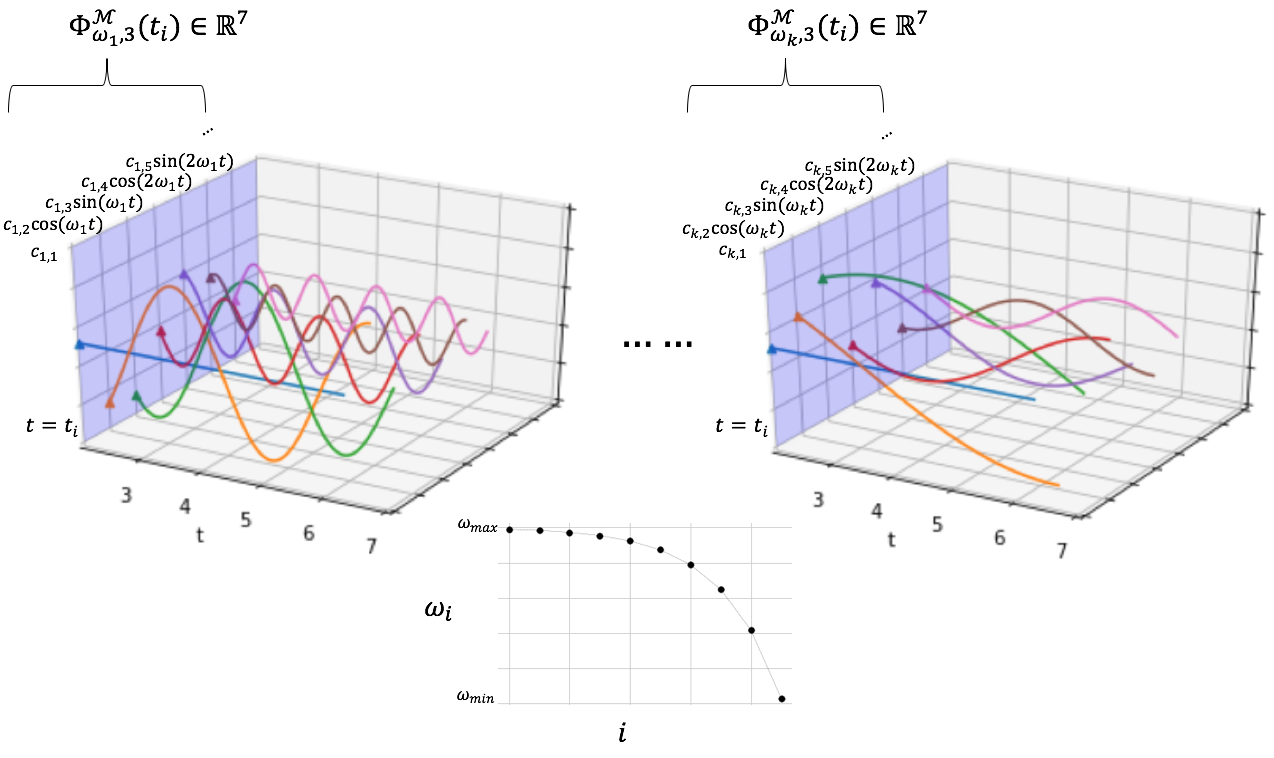} \\ \hline 
    \end{tabular}
    \caption{Sketched visual illustration of the proposed Bochner and Mercer time embedding ($\Phi^{\mathcal{B}}_d(t)$ and $\Phi^{\mathcal{M}}_{\omega,d}(t)$) for a specific $t=t_i$ with $d=3$. In right panel the scale of sine and cosine waves decreases as their frequency gets larger, which is a common phenomenon for Fourier series.} 
    \label{tab:visualization}
\end{table}

\section{Time-event Interaction}
\label{sec:interaction}
Learning time-event interaction is crucial for continuous-time event sequence prediction. After embedding time span into finite-dimensional vector spaces, we are able to directly model interactions using time and event embeddings. It is necessary to first project the time and event representations onto the same space.
For an event sequence $\big\{ (e_1, t_1), \ldots, (e_q, t_q)\big \}$ we concatenate the event and time representations into $[\mathbf{Z}, \mathbf{Z}_T]$ where $\mathbf{Z}=\big[Z_1,\ldots,Z_q\big]$, $\mathbf{Z}_T = \big[\Phi(t_1), \ldots, \Phi(t_q)\big]$ and project them into the query, key and value spaces. For instance, to consider only linear combinations of event and time representations in query space, we can simply use $\mathbf{Q}=[\mathbf{Z}, \mathbf{Z}_T]\mathbf{W}_0 + b_0$. To capture non-linear relations hierarchically, we may consider using multilayer perceptrons (MLP) with activation functions, such as 
\[\mathbf{Q} = \text{ReLU}\big([\mathbf{Z}, \mathbf{Z}_T]\mathbf{W}_0 + b_0\big)\mathbf{W}_1+b_1,\]
where $\text{ReLU}(.)$ is the rectified linear unit. Residual blocks can also be added to propagate useful lower-level information to final output.
When predicting the next time-dependent event $(e_{q+1}, t_{q+1})$, to take account of the time lag between each event in input sequence and target event we let $\tilde{t}_i = t_{q+1} - t_{i}, i=1,\ldots,q$ and use $\Phi(\tilde{t}_i)$ as time representations. This does not change the relative time difference between input events, i.e. $\tilde{t}_i - \tilde{t}_j = t_i - t_j$ for $i,j=1,\ldots,q$, and now the attention weights and prediction becomes a function of next occurrence time.

\section{Experiment and Result}
\label{sec:experiment}
We evaluate the performance of the proposed time embedding methods with self-attention on several real-world datasets from various domains. The experiemnts aim at quantitatively evaluating the performance of the four time embedding methods, and comparing them with baseline models. 
\subsection{Data Sets}




\begin{itemize}
    \item \textbf{Stack Overflow}\footnote{https://archive.org/details/stackexchange} dataset records user's history awarded badges in a question-answering website. The task is to predict the next badge the user receives, as a classification task.
    
    \item \textbf{MovieLens}\footnote{https://grouplens.org/datasets/movielens/1m/} is a public dataset consists of movie rating for benchmarking recommendations algorithms \cite{Harper:2015:MDH:2866565.2827872}. 
    The task is to predict the next movie that the user rates for recommendation.
    
    \item \textbf{Walmart.com dataset} is obtained from Walmart's online e-commerce platform in the U.S\footnote{https://www.walmart.com}. It contains the session-based search, view, add-to-cart and transaction information with timestamps for each action from selected users. The task is to predict the next-view item for recommendation. Details for all datasets are provided in supplemnetary meterial.
\end{itemize}

\textbf{Data preparation} - For fair comparisons with the baselines, on the \textsl{MovieLens} dataset we follow the same prepossessing steps mentioned in \cite{kang2018self}. For users who rated at least three movies, we use their second last rating for validation and their last rated movie for testing. On the \textsl{stack overflow} dataset we use the same filtering procedures described in \cite{li2017time} and randomly split the dataset on users into training (80\%), validation (10\%) and test (10\%). On the \textsl{Walmart.com} dataset we filter out users with less than ten activities and products that interacted with less than five users. The training, validation and test data are splited based on session starting time chronically.

\subsection{Baselines and Model configurations}
We compare the proposed approach with LSTM, the time-aware RNN model (\textsl{TimeJoint}) \cite{li2017time} and recurrent marked temporal point process model (\textsl{RMTPP}) \cite{du2016recurrent} on the \textsl{Stack Overflow} dataset. We point out that the two later approaches also utilize time information. For the above three models, we reuse the optimal model configurations and metrics (classification \emph{accuracy}) reported in \cite{li2017time} for the same \textsl{Stack Overflow} dataset. 

For the recommendation tasks on \textsl{MovieLens} dataset, we choose the seminal session-based RNN recommendation model (\textsl{GRU4Rec}) \cite{hidasi2015session}, convolutional sequence embedding method (\textsl{Caser}) \cite{tang2018personalized} and translation-based recommendation model (\textsl{TransRec}) \cite{kang2018self} as baselines. These position-aware sequential models have been shown to achieve cutting-edge performances on the same \textsl{MovieLens} dataset \cite{kang2018self}. We also reuse the metrics - top K hitting rate (\emph{Hit@K}) and normalized discounted cumulative gain (\emph{NDCG@K}), as well as the optimal model configurations reported in \cite{kang2018self}.  

On the \textsl{Walmart.com} dataset, other than \textsl{GRU4Rec} and \textsl{TransRec}, we compare with an attention-based RNN model \textsl{RNN+attn}. The hyper-parameters of the baselines are tuned for optimal performances according to the \emph{Hit@10} metric on the validation dataset. The outcomes are provided in Table \ref{tab:moivelens}.

As for the proposed time embedding methods, we experimented on the Bochner time embedding with the \textsl{reparameterization trick} using normal distribution (\textsl{Bochner Normal}), the parametric inverse CDF transformation (\textsl{Bochner Inv CDF}) with MLP, MLP + residual block, masked autoregressive flow (\textsl{MAF}) \cite{papamakarios2017masked} and non-volume preserving transformations (\textsl{NVP}) \cite{dinh2016density}, the non-parametric inverse CDF transformation (\textsl{Bochner Non-para}), as well as the Mercer time embedding. For the purpose of \emph{ablation study}, we compare with the original positional encoding self-attention (\textsl{PosEnc}) for all tasks (Table \ref{tab:moivelens}). We use $d=100$ for both Bochner and Mercer time embedding, with the sensitivity analysis on time embedding dimensions provided in appendix. We treat the dimension of Fourier basis $k$ for Mercer time embedding as hyper-parameter, and select from $\{1,5,10,15,20,25,30\}$ according to the validation \emph{Hit@10} as well. When reporting the results in Table \ref{tab:moivelens}, we mark the model configuration that leads to the optimal validation performance for each of our time embedding methods. Other configurations and training details are provided in appendix.


\subsection{Experimental results}

\begin{table}[tbh]
\resizebox{14cm}{!}
{
\footnotesize
    \centering
    \begin{tabular}{c c c c | p{1.1cm} p{1.1cm} p{1.3cm} p{1.4cm} p{1.4cm}}
    \hline \hline 
        \multicolumn{8}{c}{Stack Overflow}\\ \hline
         \hline 
         \textbf{Method} & LSTM & TimeJoint & RMTPP & PosEnc & \textbf{Bochner Normal} & \textbf{Bochner Inv CDF} & \textbf{Bochner Non-para} & \textbf{Mercer} \\
         
         \textbf{Accuracy} & 46.03(.21) & 46.30(.23) & 46.23(.24) & 44.03(.33) & 44.89(.46) & 44.67(.38) & 46.27(0.29) & \textbf{46.83(0.20)} \\ \hdashline
         config &  &  &  &  &  & \textsl{NVP} &  & $k=10$ \\ \hline
         \hline
        \multicolumn{8}{c}{MovieLens-1m}\\ \hline
        \hline 
         \textbf{Method} & GRU4Rec & Caser & TransRec & - & - & - & - & - \\ 
         \textbf{Hit@10} & 75.01(.25) & 78.86(.22)& 64.15(.27) & 82.45(.31) & 81.60(.69) & 82.52(.36) & 82.86(.22) & \textbf{82.92 (.17)}\\ 
         \textbf{NDCG@10} &55.13(.14) & 55.38(.15)& 39.72(.16) & 59.05(.14) & 59.47(.56) & 60.80(.47) & 60.83(.15) & \textbf{61.67 (.11)}\\ \hdashline
         config &  &  &  &  &  & \textsl{MAF} &  & $k=5$ \\ \hline
         \hline

         \multicolumn{8}{c}{Walmart.com data}\\ \hline
         \hline
         \textbf{Method} & GRU4Rec & RNN+attn & TransRec & - & - & - & - & - \\ 
         
         \textbf{Hit@5} & 4.12(.19) & 5.90(.17)& 7.03(.15) & 8.63(.16) & 4.27(.91) & 9.04(.31) & 9.25(.15) & \textbf{10.92(.13)}\\
         \textbf{NDCG@5} & 4.03(.20) & 4.66(.17) & 5.62(.17) & 6.92(.14) & 4.06(.94) & 7.27(.26) & 7.34(.12) & \textbf{8.90(.11)} \\
         \textbf{Hit@10} & 6.71(.50) & 9.03(.44) & 10.38(.41) & 12.49(.38) & 7.66(.92) & 12.77(.65) & 13.16(.41) & \textbf{14.94(.31)}\\ 
         \textbf{NDCG@10} & 4.97(.31) & 7.36(.26) & 8.72(.26) & 10.84(.26) & 6.02(.99) & 10.95(.74) & 11.36(.27)  & \textbf{12.81(.22)} \\  \hdashline
         config &  &  &  &  &  & \textsl{MAF} &  & $k=25$ \\ \hline
    \end{tabular}
}
    \caption{Performance metrics for the proposed apporach and baseline models. All results are converted to percentage by multiplying by 100, and the standard deviations computed over ten runs are given in the parenthesis. The proposed methods and the best outcomes are highlighted in bold font. The \emph{config} rows give the optimal model configuration for \textsl{Bochner Inv CDF} (among using MLP, MLP + redisual block, \textsl{MAF} and \textsl{NVP} as CDF learning method) and \textsl{Mercer} (among $k=1,5,\ldots,30$).}  
    \label{tab:moivelens}
\end{table}

\begin{figure*}[hbt]
    \centering
    \begin{subfigure}[t]{0.4\textwidth}
        \centering
        \includegraphics[scale = 0.34]{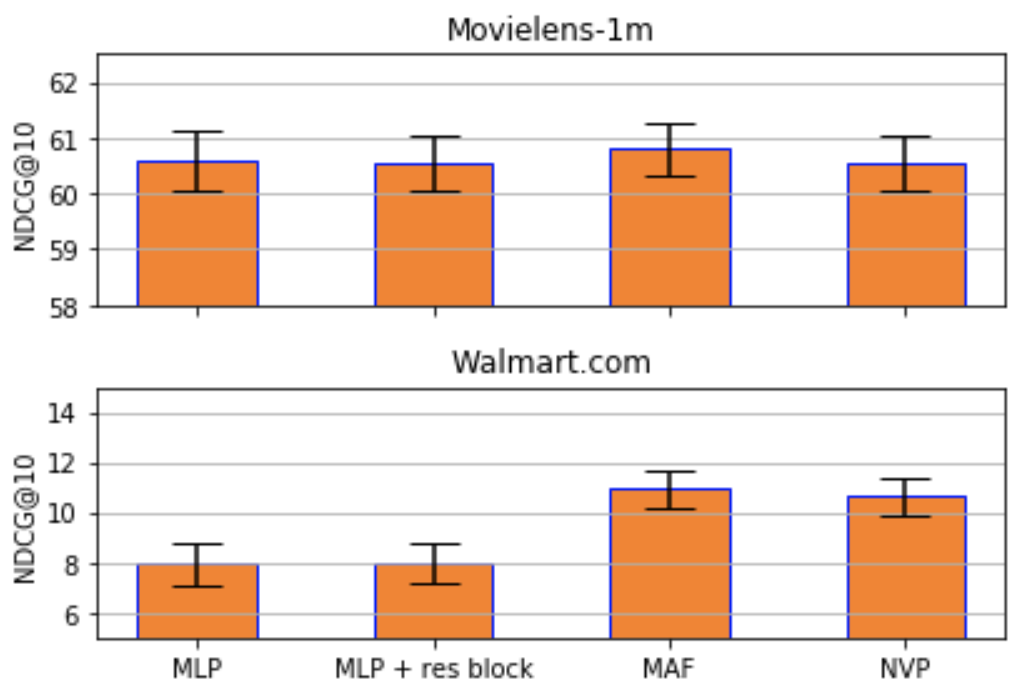}
    \end{subfigure}%
    ~ 
    \begin{subfigure}[t]{0.6\textwidth}
        \centering
        \includegraphics[scale = 0.20]{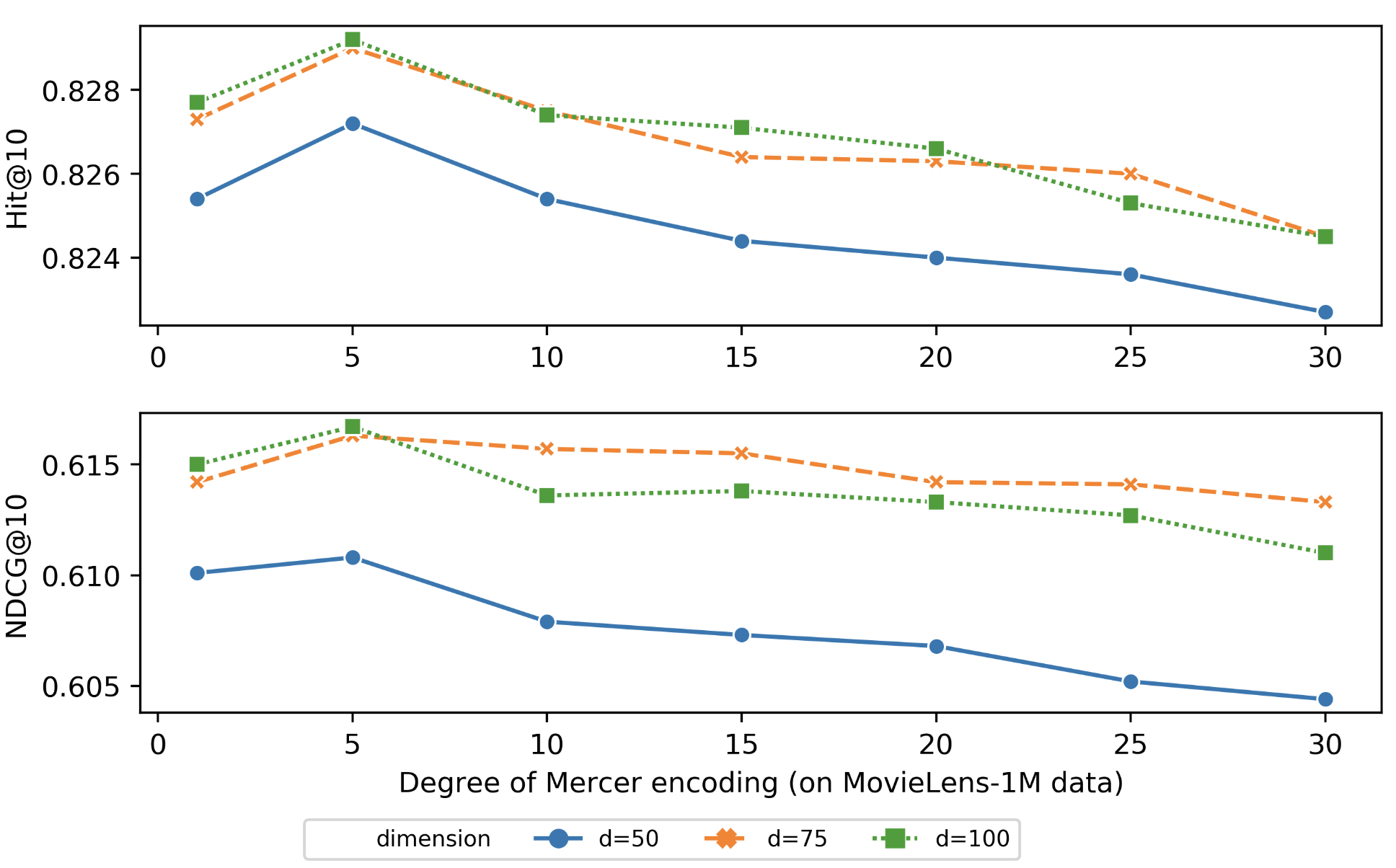}
    \end{subfigure}
    
    \caption{(a). We show the results of \textsl{Bochner Inv CDF} on the \textsl{Movielens} and \textsl{Walmart.com} dataset with different distributional learning methods. (b). The sensitivity analysis on Mercer time encoding on the \textsl{Movielens} dataset by varying the degree of Fourier basis $k$ under different dimension $d$.}
    \label{fig:bochner-methods}
\end{figure*}

We observe in Table \ref{tab:moivelens} that the proposed time embedding with self-attention compares favorably to baseline models on all three datasets. For the \textsl{Stack Overflow} and \textsl{Walmart.com} dataset, \textsl{Mercer} time embedding achieves best performances, and on \textsl{MovieLens} dataset the \textsl{Bochner Non-para} outperforms the remaining methods. The results suggest the effectiveness of the functional time representation, and the comparison with positional encoding suggests that time embedding are more suitable for continuous-time event sequence modelling. On the other hand, it appears that \textsl{Bochner Normal} and \textsl{Bochner Inv CDF} has higher variances, which might be caused by their need for sampling steps during the training process. Otherwise, \textsl{Bochner Inv CDF} has comparable performances to \textsl{Bochner Non-para} across all three datasets. In general, we observe better performances from \textsl{Bochner Non-para} time embedding and Mercer time embedding. Specifically, with the tuned Fourier basis degree $k$, Mercer's method consistently outperforms others across all tasks. While $d$, the dimension of time embedding, controls how well the bandwidth of [$\omega_{\min},\omega_{\max}$] is covered, $k$ controls the degree of freedom for the Fourier basis under each frequency. When $d$ is fixed, larger $k$ may lead to overfitting issue for the time kernels under certain frequencies, which is confirmed by the sensitivity analysis on $k$ provided in Figure \ref{fig:bochner-methods}b.

\begin{figure*}[hbt]
    \centering
    \begin{subfigure}[t]{0.3\textwidth}
        \centering
        \includegraphics[scale = 0.25]{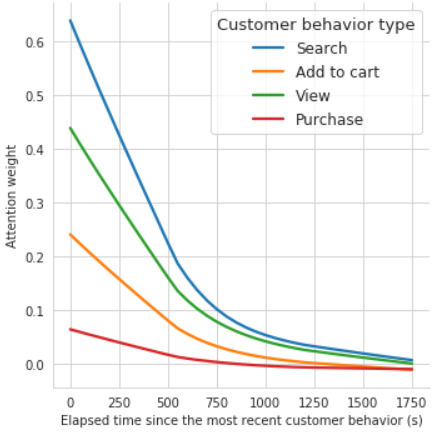}
        \caption{The temporal patterns in average attention weight decay on the last interacted product after different user actions, as time elapsed.}
        \label{fig:time-event1}
    \end{subfigure}%
    ~ 
    \begin{subfigure}[t]{0.3\textwidth}
        \centering
        \includegraphics[scale = 0.25]{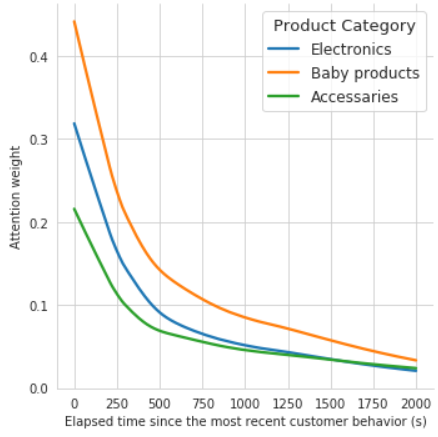}
        \caption{The temporal patterns in average attention weight decay on the last viewed product from different departments, as time elapsed.}
        \label{fig:time-event2}
    \end{subfigure}
    ~
    \begin{subfigure}[t]{0.3\textwidth}
        \centering
        \includegraphics[scale = 0.25]{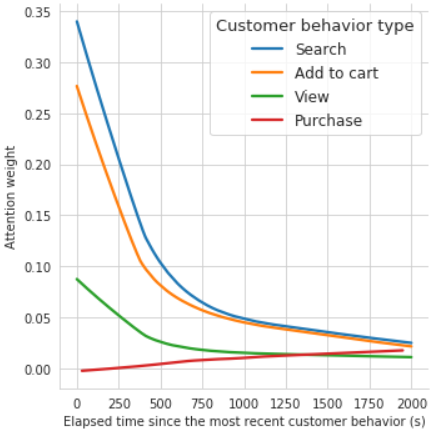}
        \caption{The prediction of future attention weight on the last interacted product as a function of time and different user actions.}
        \label{fig:time-event3}
    \end{subfigure}
    \caption{Temporal patterns and time-event interactions captured by time and event representations on the Walmart.com dataset.}
    \label{fig:time-event}
\end{figure*}

In Figure \ref{fig:time-event}, we visualize the average attention weights across the whole population as functions of time and user action or product department on the Walmart.com dataset, to demonstrate some of the useful temporal patterns captured by the Mercer time embedding. For instance, Figure \ref{fig:time-event1} shows that when recommending the next product, the model learns to put higher attention weights on the last searched products over time. Similarly, the patterns in Figure \ref{fig:time-event2} indicate that the model captures the signal that customers often have prolonged or recurrent attentions on \textsl{baby products} when compared with \textsl{electronics} and \textsl{accessories}. Interestingly, when predicting the attention weights by using future time points as input (Figure \ref{fig:time-event3}), we see our model predicts that the users almost completely lose attention on their most recent purchased products (which is reasonable), and after a more extended period none of the previously interacted products matters anymore. 


\textbf{Discussion}. By employing state-of-the-art CDF learning methods, \textsl{Bochner Inv CDF} achieves better performances than positional encoding and other baselines on Movlielens and Walmart.com dataset (Figure \ref{fig:bochner-methods}a). This suggests the importance of having higher model complexity for learning the $p(\omega)$ in Bochner's Thm, and also explains why \textsl{Bochner Normal} fails since normal distribution has limited capacity in capturing complicated distributional signals. On the other hand, \textsl{Bochner Non-para} is actually the special case of \textsl{Mercer}'s method with $k=1$ and no intercept. While Bochner's methods originate from random feature sampling, Mercer's method grounds in functional basis expansion. In practice, we may expect Mercer's method to give more stable performances since it does not rely on distributional learning and sampling. However, with advancements in Bayesian deep learning and probabilistic computation, we may also expect \textsl{Bochner Inv CDF} to work appropriately with suitable distribution learning models, which we leave to future work.

\section{Conlusion}
\label{sec:conclusion}
We propose a set of time embedding methods for functional time representation learning, and demonstrate their effectiveness when using with self-attention in continuous-time event sequence prediction. The proposed methods come with sound theoretical justifications, and not only do they reveal temporal patterns,  but they also capture time-event interactions. The proposed time embedding methods are thoroughly examined by experiments using real-world datasets, and we find Mercer time embedding and Bochner time embedding with non-parametric inverse CDF transformation giving superior performances. We point out that the proposed methods extend to general time representation learning, and we will explore adapting our proposed techniques to other settings such as temporal graph representation learning and reinforcement learning in the our future work.

\bibliographystyle{abbrvnat}
\bibliography{references}

\begin{thebibliography}{27}
\providecommand{\natexlab}[1]{#1}
\providecommand{\url}[1]{\texttt{#1}}
\expandafter\ifx\csname urlstyle\endcsname\relax
  \providecommand{\doi}[1]{doi: #1}\else
  \providecommand{\doi}{doi: \begingroup \urlstyle{rm}\Url}\fi

\bibitem[Bahdanau et~al.(2014)Bahdanau, Cho, and Bengio]{bahdanau2014neural}
D.~Bahdanau, K.~Cho, and Y.~Bengio.
\newblock Neural machine translation by jointly learning to align and
  translate.
\newblock \emph{arXiv preprint arXiv:1409.0473}, 2014.

\bibitem[Bengio et~al.(2013)Bengio, Courville, and
  Vincent]{bengio2013representation}
Y.~Bengio, A.~Courville, and P.~Vincent.
\newblock Representation learning: A review and new perspectives.
\newblock \emph{IEEE transactions on pattern analysis and machine
  intelligence}, 35\penalty0 (8):\penalty0 1798--1828, 2013.

\bibitem[Chen et~al.(2017)Chen, Zhang, Xiao, Nie, Shao, Liu, and
  Chua]{chen2017sca}
L.~Chen, H.~Zhang, J.~Xiao, L.~Nie, J.~Shao, W.~Liu, and T.-S. Chua.
\newblock Sca-cnn: Spatial and channel-wise attention in convolutional networks
  for image captioning.
\newblock In \emph{Proceedings of the IEEE conference on computer vision and
  pattern recognition}, pages 5659--5667, 2017.

\bibitem[Chorowski et~al.(2015)Chorowski, Bahdanau, Serdyuk, Cho, and
  Bengio]{chorowski2015attention}
J.~K. Chorowski, D.~Bahdanau, D.~Serdyuk, K.~Cho, and Y.~Bengio.
\newblock Attention-based models for speech recognition.
\newblock In \emph{Advances in neural information processing systems}, pages
  577--585, 2015.

\bibitem[Dinh et~al.(2016)Dinh, Sohl-Dickstein, and Bengio]{dinh2016density}
L.~Dinh, J.~Sohl-Dickstein, and S.~Bengio.
\newblock Density estimation using real nvp.
\newblock \emph{arXiv preprint arXiv:1605.08803}, 2016.

\bibitem[Du et~al.(2016)Du, Dai, Trivedi, Upadhyay, Gomez-Rodriguez, and
  Song]{du2016recurrent}
N.~Du, H.~Dai, R.~Trivedi, U.~Upadhyay, M.~Gomez-Rodriguez, and L.~Song.
\newblock Recurrent marked temporal point processes: Embedding event history to
  vector.
\newblock In \emph{Proceedings of the 22nd ACM SIGKDD International Conference
  on Knowledge Discovery and Data Mining}, pages 1555--1564. ACM, 2016.

\bibitem[Harper and Konstan(2015)]{Harper:2015:MDH:2866565.2827872}
F.~M. Harper and J.~A. Konstan.
\newblock The movielens datasets: History and context.
\newblock \emph{ACM Trans. Interact. Intell. Syst.}, 5\penalty0 (4):\penalty0
  19:1--19:19, Dec. 2015.
\newblock ISSN 2160-6455.
\newblock \doi{10.1145/2827872}.
\newblock URL \url{http://doi.acm.org/10.1145/2827872}.

\bibitem[Hidasi et~al.(2015)Hidasi, Karatzoglou, Baltrunas, and
  Tikk]{hidasi2015session}
B.~Hidasi, A.~Karatzoglou, L.~Baltrunas, and D.~Tikk.
\newblock Session-based recommendations with recurrent neural networks.
\newblock \emph{arXiv preprint arXiv:1511.06939}, 2015.

\bibitem[Jackson(1930)]{jackson1930theory}
D.~Jackson.
\newblock \emph{The theory of approximation}, volume~11.
\newblock American Mathematical Soc., 1930.

\bibitem[Kang and McAuley(2018)]{kang2018self}
W.-C. Kang and J.~McAuley.
\newblock Self-attentive sequential recommendation.
\newblock In \emph{2018 IEEE International Conference on Data Mining (ICDM)},
  pages 197--206. IEEE, 2018.

\bibitem[Kingma and Welling(2013)]{kingma2013auto}
D.~P. Kingma and M.~Welling.
\newblock Auto-encoding variational bayes.
\newblock \emph{arXiv preprint arXiv:1312.6114}, 2013.

\bibitem[Li et~al.(2017)Li, Du, and Bengio]{li2017time}
Y.~Li, N.~Du, and S.~Bengio.
\newblock Time-dependent representation for neural event sequence prediction.
\newblock \emph{arXiv preprint arXiv:1708.00065}, 2017.

\bibitem[Loomis(2013)]{loomis2013introduction}
L.~H. Loomis.
\newblock \emph{Introduction to abstract harmonic analysis}.
\newblock Courier Corporation, 2013.

\bibitem[Mei and Eisner(2017)]{mei2017neural}
H.~Mei and J.~M. Eisner.
\newblock The neural hawkes process: A neurally self-modulating multivariate
  point process.
\newblock In \emph{Advances in Neural Information Processing Systems}, pages
  6754--6764, 2017.

\bibitem[Mercer(1909)]{mercer1909xvi}
J.~Mercer.
\newblock Xvi. functions of positive and negative type, and their connection
  the theory of integral equations.
\newblock \emph{Philosophical transactions of the royal society of London.
  Series A, containing papers of a mathematical or physical character},
  209\penalty0 (441-458):\penalty0 415--446, 1909.

\bibitem[Papamakarios et~al.(2017)Papamakarios, Pavlakou, and
  Murray]{papamakarios2017masked}
G.~Papamakarios, T.~Pavlakou, and I.~Murray.
\newblock Masked autoregressive flow for density estimation.
\newblock In \emph{Advances in Neural Information Processing Systems}, pages
  2338--2347, 2017.

\bibitem[Rahimi and Recht(2008)]{rahimi2008random}
A.~Rahimi and B.~Recht.
\newblock Random features for large-scale kernel machines.
\newblock In \emph{Advances in neural information processing systems}, pages
  1177--1184, 2008.

\bibitem[Rezende and Mohamed(2015)]{rezende2015variational}
D.~J. Rezende and S.~Mohamed.
\newblock Variational inference with normalizing flows.
\newblock \emph{arXiv preprint arXiv:1505.05770}, 2015.

\bibitem[Tang and Wang(2018)]{tang2018personalized}
J.~Tang and K.~Wang.
\newblock Personalized top-n sequential recommendation via convolutional
  sequence embedding.
\newblock In \emph{Proceedings of the Eleventh ACM International Conference on
  Web Search and Data Mining}, pages 565--573. ACM, 2018.

\bibitem[Vaswani et~al.(2017)Vaswani, Shazeer, Parmar, Uszkoreit, Jones, Gomez,
  Kaiser, and Polosukhin]{vaswani2017attention}
A.~Vaswani, N.~Shazeer, N.~Parmar, J.~Uszkoreit, L.~Jones, A.~N. Gomez,
  {\L}.~Kaiser, and I.~Polosukhin.
\newblock Attention is all you need.
\newblock In \emph{Advances in neural information processing systems}, pages
  5998--6008, 2017.

\bibitem[Widom(1964)]{widom1964asymptotic}
H.~Widom.
\newblock Asymptotic behavior of the eigenvalues of certain integral equations.
  ii.
\newblock \emph{Archive for Rational Mechanics and Analysis}, 17\penalty0
  (3):\penalty0 215--229, 1964.

\bibitem[Xiao et~al.(2017{\natexlab{a}})Xiao, Yan, Farajtabar, Song, Yang, and
  Zha]{xiao2017joint}
S.~Xiao, J.~Yan, M.~Farajtabar, L.~Song, X.~Yang, and H.~Zha.
\newblock Joint modeling of event sequence and time series with attentional
  twin recurrent neural networks.
\newblock \emph{arXiv preprint arXiv:1703.08524}, 2017{\natexlab{a}}.

\bibitem[Xiao et~al.(2017{\natexlab{b}})Xiao, Yan, Yang, Zha, and
  Chu]{xiao2017modeling}
S.~Xiao, J.~Yan, X.~Yang, H.~Zha, and S.~M. Chu.
\newblock Modeling the intensity function of point process via recurrent neural
  networks.
\newblock In \emph{Thirty-First AAAI Conference on Artificial Intelligence},
  2017{\natexlab{b}}.

\bibitem[Xu et~al.(2019)Xu, Ruan, Korpeoglu, Kumar, and Achan]{xu2019context}
D.~Xu, C.~Ruan, E.~Korpeoglu, S.~Kumar, and K.~Achan.
\newblock Context-aware dual representation learning for complementary products
  recommendation.
\newblock \emph{arXiv preprint arXiv:1904.12574v2}, 2019.

\bibitem[Xu et~al.(2015)Xu, Ba, Kiros, Cho, Courville, Salakhutdinov, Zemel,
  and Bengio]{xu2015show}
K.~Xu, J.~Ba, R.~Kiros, K.~Cho, A.~Courville, R.~Salakhutdinov, R.~Zemel, and
  Y.~Bengio.
\newblock Show, attend and tell: Neural image caption generation with visual
  attention.
\newblock \emph{arXiv preprint arXiv:1502.03044}, 2015.

\bibitem[Zhao et~al.(2015)Zhao, Erdogdu, He, Rajaraman, and
  Leskovec]{zhao2015seismic}
Q.~Zhao, M.~A. Erdogdu, H.~Y. He, A.~Rajaraman, and J.~Leskovec.
\newblock Seismic: A self-exciting point process model for predicting tweet
  popularity.
\newblock In \emph{Proceedings of the 21th ACM SIGKDD International Conference
  on Knowledge Discovery and Data Mining}, pages 1513--1522. ACM, 2015.

\bibitem[Zhu et~al.(2017)Zhu, Li, Liao, Wang, Guan, Liu, and Cai]{zhu2017next}
Y.~Zhu, H.~Li, Y.~Liao, B.~Wang, Z.~Guan, H.~Liu, and D.~Cai.
\newblock What to do next: Modeling user behaviors by time-lstm.
\newblock In \emph{IJCAI}, pages 3602--3608, 2017.

\end{thebibliography}

\newpage 
\appendix 
\section{Appendix}
\subsection{Proof of Claim 1}


\begin{proof}
Define the score $S(t_1,t_2)=\Phi^{\mathcal{B}}_d(t_1)^{'}\Phi^{\mathcal{B}}_d(t_2)$. The goal is to derive a uniform upper bound for $s(t_1,t_2) - \mathcal{K}(t_1,t_2)$. By assumption $S(t_1,t_2)$ is an unbiased estimator for $\mathcal{K}(t_1,t_2)$, i.e. $E[S(t_1,t_2)] = \mathcal{K}(t_1,t_2)$. Due to the translation-invariant property of $S$ and $\mathcal{K}$, we let $\Delta(t) \equiv s(t_1,t_2) - \mathcal{K}(t_1,t_2)$, where $t \equiv t_1 - t_2$ for all $t_1,t_2 \in [0,t_{\max}]$. Also we define $s(t_1 - t_2 ):=S(t_1,t_2)$. Therefore $t \in [-t_{\max}, t_{\max}]$, and we use $t \in \tilde{T}$ as the shorthand notation. The LHS in (1) now becomes $\text{Pr}\big(\sup_{t \in \tilde{T}}|\Delta(t)| \geq \epsilon \big)$.

Note that $\tilde{T} \subseteq \cup_{i=0}^{N-1} T_i$ with $T_i = \big[-t_{\max} + \frac{2it_{\max}}{N}, -t_{\max} + \frac{2(i+1)t_{\max}}{N}\big]$ for $i=1,\ldots,N$. So $\cup_{i=0}^{N-1} T_i$ is a finite cover of $\tilde{T}$. Define $t_i = -t_{\max} + \frac{(2i+1)t_{\max}}{N}$, then for any $t \in T_i$, $ i=1,\ldots,N$ we have 
\begin{equation}
\label{eqn:appn2}
    \begin{split}
        \big|\Delta(t)\big| &= \big|\Delta(t) - \Delta(t_i) + \Delta(t_i)\big| \\
        & \leq \big|\Delta(t) - \Delta(t_i)\big| + \big|\Delta(t_i)\big| \\
        & \leq L_{\Delta}\big|t - t_i\big| + \big|\Delta(t_i)\big| \\
        & \leq L_{\Delta}\frac{2t_{\max}}{N} + \big|\Delta(t_i)\big|,
    \end{split}
\end{equation}
where $L_{\Delta} = \max_{t \in \tilde{T}} \big\|\nabla \Delta(t)\big\|$ (since $\Delta$ is differentiable) with the maximum achieved at $t^{*}$. So we may bound the two events separately. 

For $|\Delta(t_i)|$ we simply notice that trigeometric functions are bounded between $[-1,1]$, and therefore $-1 \leq \Phi^{\mathcal{B}}_d(t_1)^{'}\Phi^{\mathcal{B}}_d(t_2) \leq 1$. The Hoeffding's inequality for bounded random variables immediately gives us:
\begin{equation*}
    \text{Pr}\big(|\Delta(t_i)| > \frac{\epsilon}{2} \big) \leq 2\exp\Big(-\frac{d\epsilon^2}{16}\Big).
\end{equation*}
So applying the Hoeffding-type union bound to the finite cover gives
\begin{equation}
\label{eqn:appn3}
    \text{Pr}\big(\cup_{i=0}^{N-1}|\Delta(t_i)| \geq \frac{\epsilon}{2}\big) \leq 2N\exp\Big(-\frac{d\epsilon^2}{16}\Big)
\end{equation}

For the other event we first apply Markov inequality and obtain:
\begin{equation}
\label{eqn:appn4}
\text{Pr}\big(L_{\Delta}\frac{2t_{\max}}{N} \geq \frac{\epsilon}{2} \big) = \text{Pr}\big(L_{\Delta} \geq \frac{\epsilon N}{4t_{\max}} \big) \leq \frac{4t_{\max}E[L_{\Delta}^2]}{\epsilon N}.
\end{equation}
Also, since $E\big[s(t_1-t_2)\big] = \psi(t_1 - t_2)$, we have
\begin{equation}
\label{eqn:appn5}
E\big[L_{\Delta}^2\big] = E\big\|\nabla s(t^*) - \nabla \psi(t^*)\big\|^2 = E\big\|\nabla s(t^*)\big\|^2 - E\big\|\nabla \psi(t^*)\big\|^2 \leq E\big\|\nabla s(t^*)\big\|^2 = \sigma_p^2,
\end{equation}
where $\sigma_p^2$ is the second momentum with respect to $p(\omega)$.

Combining (\ref{eqn:appn3}), (\ref{eqn:appn4}) and (\ref{eqn:appn3}) gives us:
\begin{equation}
\label{eqn:appn6}
    \text{Pr}\Big(\sup_{t \in \tilde{T}}|\Delta(t)| \geq \epsilon \Big) \leq 2N\exp\Big(-\frac{d\epsilon^2}{16}\Big) + \frac{4t_{\max}\sigma_p^2}{\epsilon N}.
\end{equation}
It is straightforward to examine that the RHS of (\ref{eqn:appn6}) is a convex function of $N$ and is minimized by $N^* = \sigma_p \sqrt{\frac{2t_{\max}}{\epsilon}} \exp\big(\frac{d \epsilon^2}{32}\big)$. Plug $N^*$ back to (\ref{eqn:appn6}) and we obtain bound stated in Claim 1.
\end{proof}

\subsection{Proof of Proposition 1}


\begin{proof}

We first define the kernel linear operator $\mathcal{T}_{\mathcal{K}}$ on $L^2(T)$ via $\mathcal{T}_{\mathcal{K}}(f)(t_1) = \int_{T} f(t_1)\mathcal{K}(t_1,t_2)d \mathbb{P}(t_2)$, where $\mathbb{P}$ is a non-negative measure over $T = [0,t_{\max}]$. For notation simplicity we do not explicitly index $\mathcal{K}$ with its frequency. The more complete statement of Mercer's Theorem is that under the conditions specified in Theorem 2, 
\begin{equation}
    \mathcal{T}_{\mathcal{K}}(\phi_j) = c_j \phi_j \quad \text{for } j=1,2, \ldots ,
\end{equation}
which leads to the representation of the kernel as $\mathcal{K}(x,z) = \sum_{i=1}^{\infty} c_i \phi_i(x) \phi_i(z)$. 

Therefore we only need to show the Fourier basis gives the eigenfunctions of the kernel linear operator $\mathcal{T}_{\mathcal{K}}$. Without loss of generality we assume the frequency is $2\pi$, i.e. $\psi$ is a even periodic function on $[-1,1]$ and extend to the real line by $\psi(t+2k)=\psi(t)$ for $t \in [-1, 1]$. So now the kernel linear operator is expressed by:
\begin{equation}
    \mathcal{T}_{\mathcal{K}}(f) (t_1) = \int_{-1}^{1} \psi(t_1 - t_2)f(t_2) dt_2.
\end{equation}
Now we show that the eigenfunctions for the kernel linear operator are given by Fourier basis. Suppose $\phi_{2j}(t) = \cos(\pi j t)$ for $j=1,2,\ldots$, we have
\begin{equation}
\begin{split}
    \mathcal{T}_{\mathcal{K}}(\phi_{2j})(t_1) &= \int_{-1}^{1} \psi(t_1 - t_2)\cos (2\pi t_2) dt_2 \\ 
    & \stackrel{(a)}{=} \int_{-1}^{1} \psi(t_3)\cos \big(2\pi j (t_1 + t_3)\big) dt_3 \quad (\text{with }t_3 = t_1 - t_2) \\ 
    & \stackrel{(b)}{=} \cos (\pi j t_1) \int_{-1}^1 \psi(t_3)\cos(\pi j t_3) dt_3 - \sin(\pi j t_1)\int_{-1}^1 \psi(t_3)\sin(\pi j t_3) dt_3 \\ 
    &\stackrel{(c)}{=} c_j \cos (\pi j t_1),
\end{split}
\end{equation}
where in $(a)$ we use a change of variable and utilize the periodic property of $\psi$ and the cosine function. In $(b)$ we apply the sum formula of trigonometric functions, and in $(c)$ we simply use the fact that $\int_{-1}^1 \phi(t_3)\sin(\pi j t_3) dt_3=0$ because $\psi$ is an even function. Similar arguments show that $\phi_{2j+1}(t) = \sin(\pi j t)$ for $j=1,2,\ldots$ are also the eigenfunctions for $\mathcal{T}_{\mathcal{K}}$. Since the Fourier basis form a complete orthonormal basis of $L^2(T)$, according to the complete Mercer's Theorem we see that the eigenfunctions of $\mathcal{K}$ are exactly given by the Fourier basis.
\end{proof}

\subsection{Fourier series under truncation}
In this part we briefly discuss the exponential decay for the eigenvalues $c_j$ and the uniform bound on approximation error for truncated Fourier series mentioned in Section \ref{sec:mercer}. Notice that Bochner's Theorem also applies to the periodic kernels ($\mathcal{K}(t_1,t_2) \equiv \psi(t_1 - t_2)$) stated in Mercer time embedding, such that 
\[\psi(t_1 - t_2) = \lambda \int e^{-i(t_1 - t_2) \omega}p(\omega),\] where $\lambda$ is the scaling constant such that $p(\omega)$ is a probability measure. It has been shown that if $\log p(\omega) \asymp -\omega^{a} - \log(\lambda)$ for some $a>1$, then there is a constant $b$ such that the Fourier coefficients satisfy: $c_j \asymp e^{-bj \log j}$ as $j \rightarrow \infty$ \cite{widom1964asymptotic}. 

As for the approximation error of truncated Fourier series, first we use $S_d(t_1 - t_2)$ to denote the partial sum of the Fourier series for $\psi(t_1 - t_2)$ up to the $d^{th}$ order. 
According to the Corollary \rom{1} in \cite{jackson1930theory}, if $\psi$ is $\ell -$\textsl{Lipschitz}, then we have the uniform convergence bound
\[\big|\psi(t_1 - t_2) - S_d(t_1-t_2)\big| \leq \frac{C \ell \log d}{d}\]
for all $t_1,t_2 \in T$ under some constant $C$.

The above classical results suggest the exponential decay of the Fourier coefficients as well as the uniform convergence property of the truncated Fourier series and further validate the Mercer time embedding.

\subsection{Flow-based distribution learning}
Here we briefly introduce the idea of constructing and sampling from an arbitrarily complex distribution from a known auxiliary distribution by a sequence of invertible transformations. It is motivated by the basic change of variable theorem, which we state below. 

\begin{figure}
    \centering
    \includegraphics[scale=0.3]{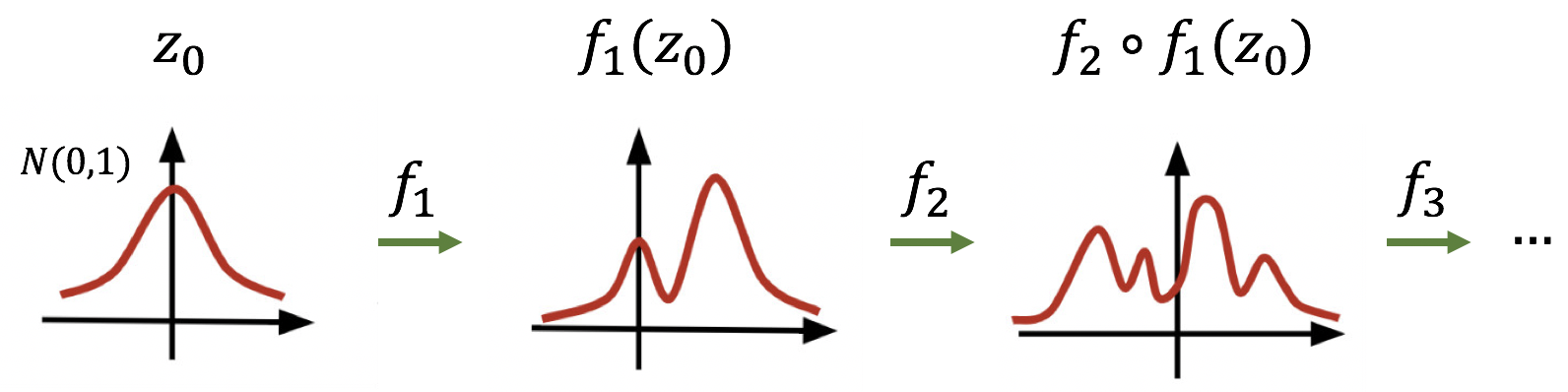}
    \caption{Illustration of the distribution transformation flow.}
    \label{fig:flow}
\end{figure}

Given an auxiliary random variable $\mathbf{z}$ following some know distribution $q(z)$, suppose another random variable $\mathbf{x}$ is constructed via a one-to-one mapping from $\mathbf{z}$: $\mathbf{x} = f(\mathbf{z})$, then the density function of $\mathbf{x}$ is given by:
\begin{equation}
    p(x) = q(z)\Big| \frac{dz}{dx} \Big| = q\big( f{\inv} (x) \big) \Big| \frac{df{\inv}}{dx} \Big|.
\end{equation}
We can parameterize the one-to-one function $f(.)$ with free parameters $\theta$ and optimize them over the observed evidence such as by maximizing the log-likelihood. By stacking a sequence of $Q$ one-to-one mappings, i.e. $\mathbf{x} = f_Q\circ f_{Q-1}\circ \ldots f_{1}(\mathbf{z})$, we can construct complicated density functions. It is easy to show by chaining that $p(x)$ is given by:
\begin{equation}
    \log p(\mathbf{x}) = \log q(\mathbf{z}) - \sum_{i=1}^Q \Big| \frac{df\inv}{dz_{i}} \Big|.
\end{equation}
A sketched graphical illustration of the concept is shown in Figure \ref{fig:flow}.

Similarity, samples from the auxiliary distribution can be transformed to the unknown target distribution in the same manner, and the transformed samples are essentially parameterized by the transformation mappings, i.e. the $g_{\theta}(\omega_i)$ in the second row of Table \ref{tab:feature_maps}.

\subsection{Dataset details}
The \textsl{Stach Overflow} dataset contains 6,000 users and 480,000 events of awarding badges. Timestamps are provided when a user is awarded a badge.  There are 22 unique badges after filtering, and the prediction of the next badge is treated as a classification task. Event sequences are generated with the same procedures described in \cite{li2017time}.

The \textsl{MovieLens} dataset, which is a benchmark for evaluating collaborative filtering algorithms, consists of 60,40 users and 3,416 movies with a total of one million ratings. The implicit feedback of rating actions characterizes user-movie interactions. Therefore the event sequence for each user is not a complete observation for their watching records. To construct event sequences from observations, we follow the same steps as described in \cite{kang2018self}, where for each user, the final rating is used for testing, the second to last rating is used for validations, and the remaining sequence is used as the input sequence.

In the \textsl{Walmart.com} dataset, there are about 72,000 users and about 1.7 million items with user-item interactions characterized by search, view, add-to-cart and transaction (purchase). The product catalog information is also available, which provides the name, brand and categories for each product. User activity records are aggregated in term of online shopping sessions. So for each user session, we construct event sequences using the same steps as the \textsl{MovieLens} dataset, in a sequence-to-sequence fashion.

\subsection{Training and model configuration}
We select the number of blocks among \{1,2,3\} and the number of attention heads among \{1,2,3,4\} for each dataset according to their validation performances. We do not experiment on using dropout or regularizations unless otherwise specified. We use the default settings for \textsl{MAF} and \textsl{NVP} provided by TensorFlow Probability
\footnote{https://www.tensorflow.org/api\_docs/python/tf/contrib/distributions/bijectors/MaskedAutoregressiveFlow} \footnote{https://www.tensorflow.org/api\_docs/python/tf/contrib/distributions/bijectors/RealNVP}
when learning the distribution for \textsl{Bochner Inv CDF}. Notice that we have not carefully tuned the \textsl{MAF} and \textsl{NVP} for \textsl{Bochner Inv CDF}, since our major focus is to show the validity of these approaches. 

\textbf{Stack Overflow} - For all the models that we implement, following the baseline settings reported in \cite{li2017time}, the hidden dimension for event representations is set to 32. The dimensions of time embeddings are also set to be 32. In each self-attention block, we concatenate time embeddings to event embeddings and project them to key, query and value spaces through linear projections, i.e.
\[
\mathbf{Q}=[\mathbf{Z}, \mathbf{Z}_T]\mathbf{W}_Q,\, \mathbf{K}=[\mathbf{Z}, \mathbf{Z}_T]\mathbf{W}_K,\,
\mathbf{V}=[\mathbf{Z}, \mathbf{Z}_T]\mathbf{W}_V,
\]
where $\mathbf{Z}$ and $\mathbf{Z}_T$ are the entity and time embeddings, $\mathbf{W}_Q$, $\mathbf{W}_K$, $\mathbf{W}_V$ are the projection matrices.
We find that using a larger hidden dimension with to many attention blocks quickly leads to over-fit in this dataset, and using the single-head self-attention gives best performances. Therefore we end up using only one self-attention block. The maximum length of the event sequence is set to be 100. For the classification problem, we feed the output sequence embeddings into a fully connected layer to predict the logits for each class and use the softmax function to compute cross-entropy loss. 

\textbf{MovieLens} - We adopt the self-attention model architecture used by the baseline models \cite{kang2018self} for fair comparisons by replacing the positional encoding with our time embedding. To be specific, the dimension for event representation is set to 50, the number of attention blocks is two and only one head is used in each attention block. To be consistent with the positional encoding self-attention baseline reported in \cite{kang2018self}, we set the dropout rate to 0.2 and the l2 regularization to 0. The maximum length of the sequence is 200, and the batch size is 128. We also adopt the shared embeddings idea for event representations \cite{kang2018self}, where we use the same set of parameters for the event embeddings layers and the final softmax layers. Finally, the cross-entropy loss with negative sampling is used to speed up the training process.

\textbf{Walmart.com dataset} -  Given the massive number of items in the dataset, we first train a shallow embeddings model to learn coarse item representations according to their context features and use those embeddings as initialization for the product representations in our model \cite{xu2019context}. The dimension of item embedding and time embedding are set to 100. Each user action (search, view, add-to-cart, transaction) is treated as a token and has a 50-dimensional vector representation that is jointly optimized as part of the model. The action embedding is concatenated to the time-event representations and together they give the time-event-action embedding. To capture time-event and time-action interactions, we first project the joint embeddings onto query, key and value spaces also with linear projections as we did on the \textsl{Stack Overflow} dataset. We find that using two attention blocks and a single head give the best results. Since the task is to predict the next-view item in the same session, we also use the cross-entropy loss with negative sampling.

During training, we apply the \emph{early stopping} where we terminate the model training if the validation performance has not increased for 10 epochs. When training models for predicting the next events, we refer to the \emph{masked self-attention} training procedure proposed in \cite{vaswani2017attention} to prevent information leak while maintaining a fast training speed.

\subsubsection{Initialization for time embedding methods} 

For the \textsl{Bochner Normal} method, we use the standard normal distribution as initial distribution in all experiments. The parametric inverse CDF function for the \textsl{Bochner Inv CDF} is carried out by a three-layer MLP under uniform initialization. For the \textsl{Bochner Non-para} method, since each $\phi_i(t) = \sin(\omega_i t)$ or $\phi_i(t) = \cos(\omega_i t)$, they have a period of $2\pi / \omega_i$. Since we would like $\phi_i$ to capture underlying temporal signals, the scale of the potential periodicity in the experimented dataset should be taken into consideration. For instance, on the \textsl{Stack Overflow} dataset, it can take days or weeks before the next event happens. Therefore, if the temporal signals were to have underlying periods, it should be on the scale from several days to several weeks. For the \textsl{Walmart.com} dataset, the next activities are often operated within minutes. Therefore the periods should range from seconds to hours. 

Therefore, in our experiments, we set the frequencies to cover a suitable range of period $[\tau_{\min}, \tau_{\max}]$ where $\tau \equiv 1/\omega$. With out loss of generality, the $\tau_{\min}$ and $\tau_{\max}$ are based on the minimum and maximum time span between consecutive events observed in data. We find that using geometric sequences that cover $[\tau_{\min}, \tau_{\max}]$ as initialization gives better results than random sampling and equal spacing sampling. To be specific, we use the set of frequencies such that their corresponding periods are given by 
\[\tau_{i} = \tau_{\min}+(\tau_{\max} - \tau_{\min})^{i/d} \quad i=1,\ldots,d.\]
Since the above argument also applies to Mercer time embedding method, we use the same initialization approach as well. 

\subsubsection{Training efficiency}

The Adam optimizer is used for all models. We set the learning rate to 0.001 and set the exponential decay rate for second moment statistics to 0.98. The training is stopped if the validation metric stops increasing in 10 consecutive epochs. We use NDCG@10 for the proprietary \textsl{Walmart.com} dataset and \textsl{MovieLens} dataset, and accuracy for \textsl{Stack Overflow} data as the monitoring metric. The final metrics are computed on the hold-out test data using model checkpoints saved during training that has the best validation performances. All models are trained in TensorFlow(1.13) on a single Nvidia V100 GPU. 

\begin{figure}
    \centering
    \includegraphics[scale=0.2]{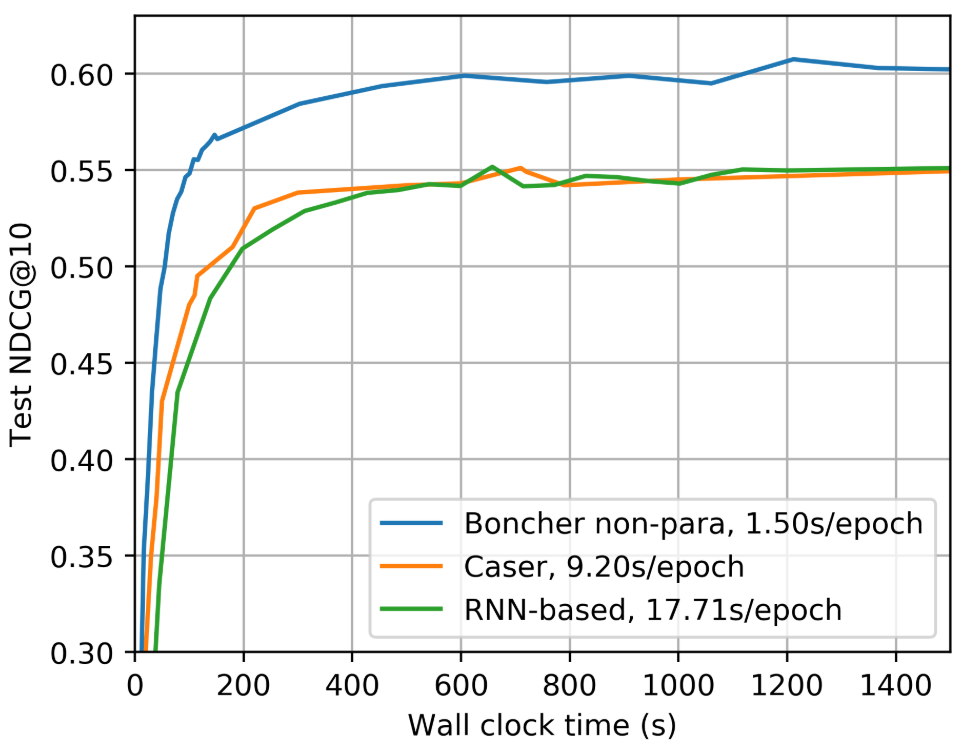}
    \caption{Training efficiency of the proposed \textsl{Bochner non-para}, the convolutional sequence embedding method \textsl{Caser} and RNN-based method on the \textsl{MovieLens} dataset. On the y-axis is the NDCG@10 on testing data.}
    \label{fig:speed_comp}
\end{figure}

The training efficiency evaluations are provided in Figure \ref{fig:speed_comp}. While it takes 9.2 seconds to train each epoch for the convolutional model \textsl{Caser} and up to 17.7 seconds for RNN-based model, it takes only 1.5 seconds for the proposed \textsl{Bochner non-para} method. Also, the test NDCG@10 reaches 0.55 within 100 seconds, while the convolutional model and RNN model reaches the same performance after 600 seconds. The training efficiency for Mercer time embedding is similar to the reported \textsl{Bochner non-para}.

\subsection{Sensitivity analysis}
\label{sec:sensitivity}

\begin{figure}[thb]
    \centering
    \includegraphics[scale=0.3]{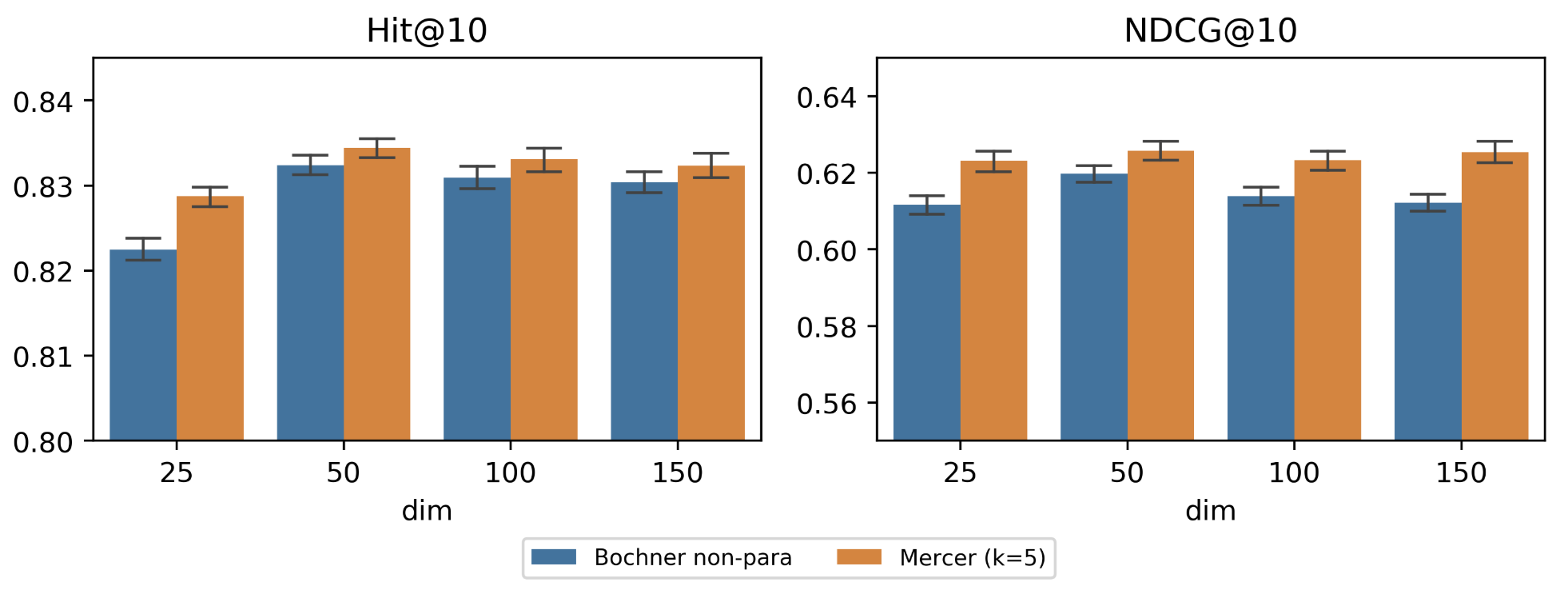}
    \caption{Sensitivity analysis for embedding dimensions on \textsl{MovieLens} data}
    \label{fig:dim_movie}
\end{figure}

\begin{figure}[thb]
    \centering
    \includegraphics[scale=0.3]{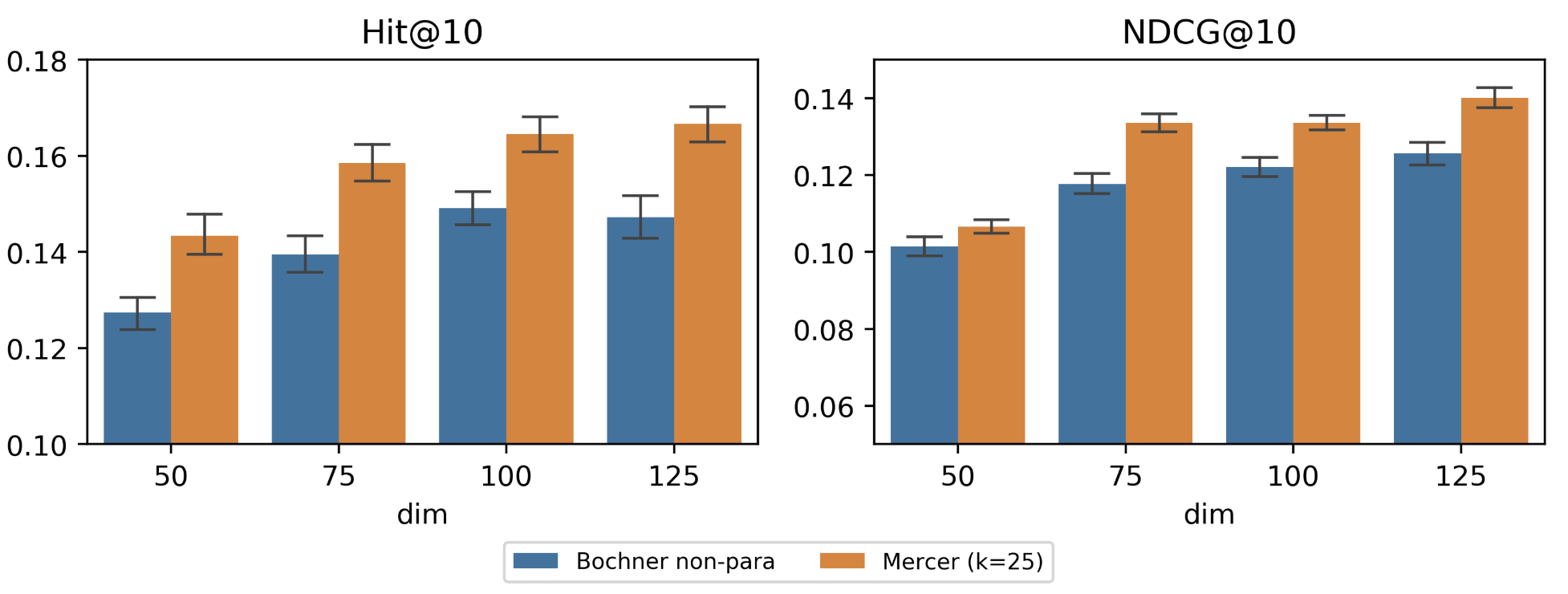}
    \caption{Sensitivity analysis for embedding dimensions on \textsl{Walmart.com} data}
    \label{fig:dim_econ}
\end{figure}

We provide sensitivity analysis on time embedding dimensions for the experiments on \textsl{MovieLens} and the proprietary \textsl{Walmart.com} dataset. We focus on the \textsl{Bochner non-para} and Mercer time embedding, which we find to have the best performances. The results are plotted in Figure \ref{fig:dim_movie} and \ref{fig:dim_econ}. For the recommendation outcomes on \textsl{MovieLens} dataset, we see that for both time embedding methods, the performances increase first and then stabilize as the dimension gets higher. On the proprietary dataset, the performance keeps increasing with larger time embedding dimensions. Firstly, the results suggest both time embedding methods have consistent performances on the two datasets. Secondly, we comment that the difference in data volumes might have caused the different trends on the two datasets. The \textsl{Walmart.com} dataset is much larger than the \textsl{MovieLens} dataset, and the temporal and time-event interaction patterns are more complicated than that of \textsl{MovieLens}. Therefore both time embedding methods keep learning with larger time embedding dimensions.

In a nutshell, the sensitivity analysis suggests that the proposed \textsl{Bochner non-para} and Mercer time embedding give stable and consistent performances on the two datasets.

\subsection{Cases study for attention weights}
\begin{figure}[tbh]
    \includegraphics[scale=0.385]{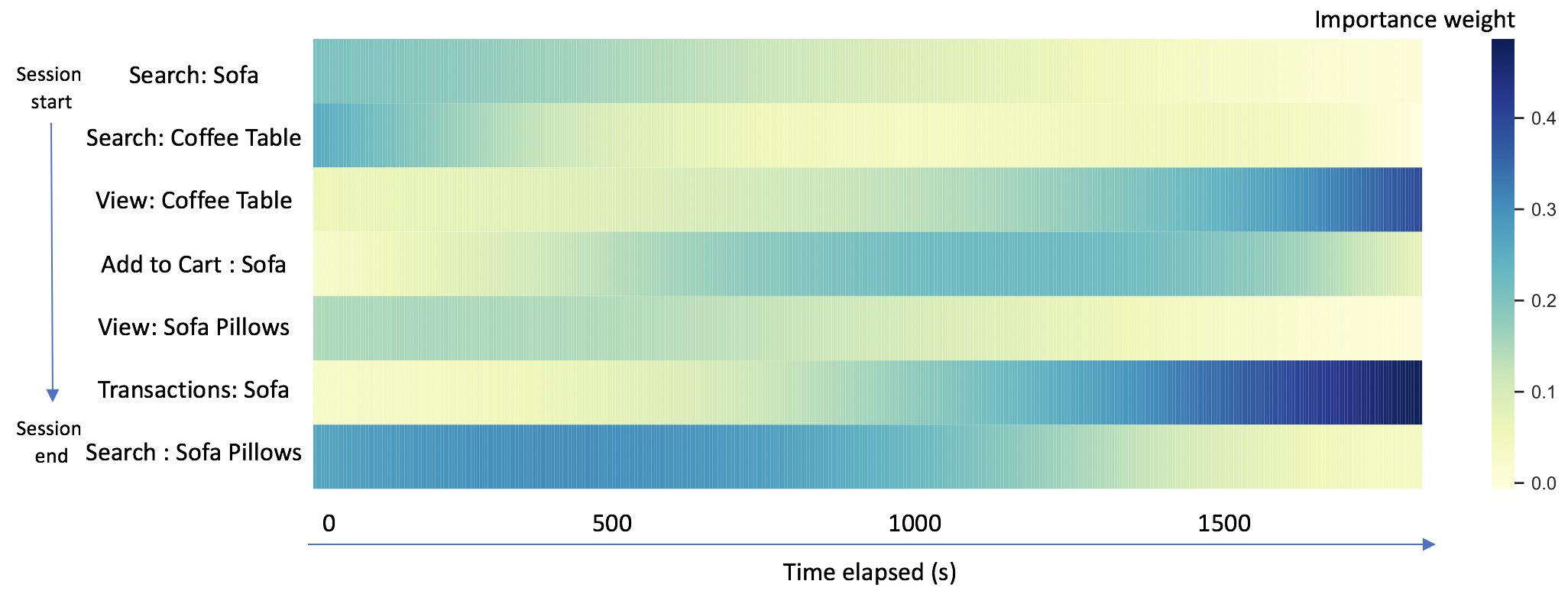}
    \caption{Dynamics of attention weights on each event-action pair with respect to the next event's occurrence time, for a real-world customer online shopping sequence in home furniture.}
    \label{fig:case_1}
\end{figure}

\begin{figure}[tbh]
    \centering
    \includegraphics[scale=0.385]{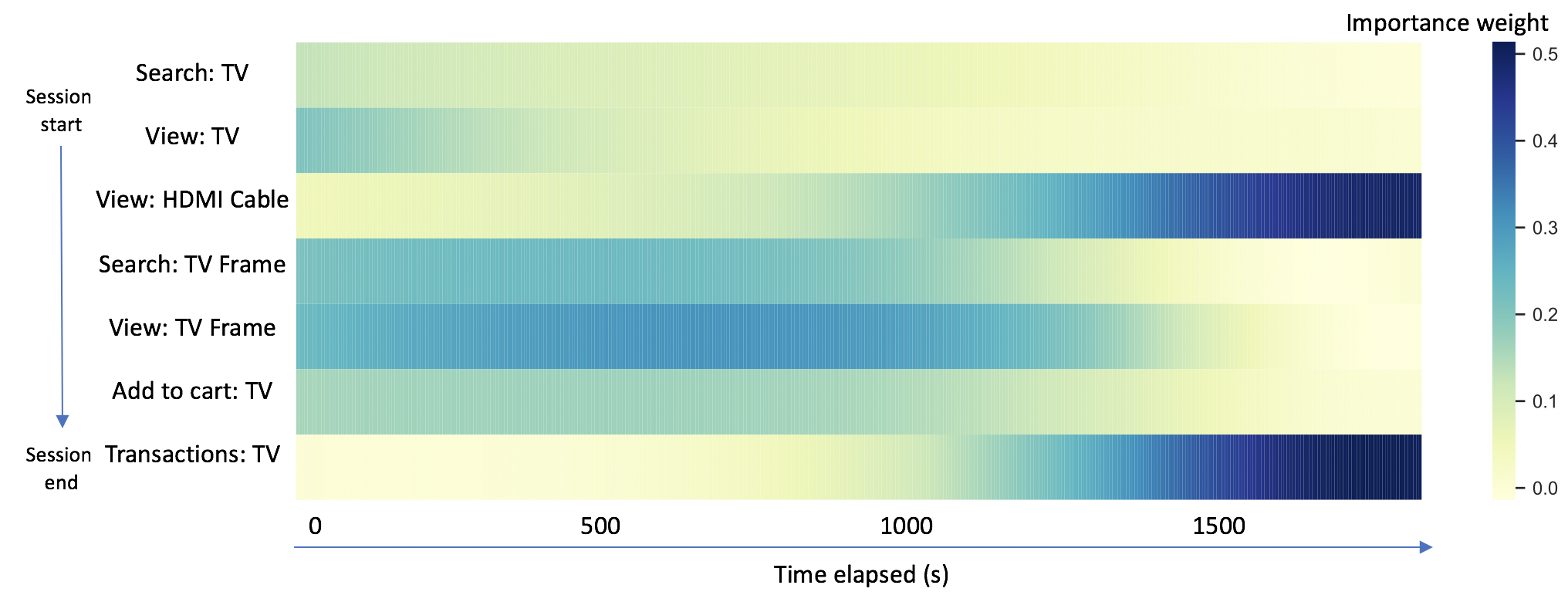}
    \caption{Dynamics of attention weights on each event-action pair with respect to the next event's occurrence time, for a real-world user online shopping sequence in TV and related electronics. }
    \label{fig:case_2}
\end{figure}

In this section, we present two user-event interaction sequences sampled from the \textsl{Walmart.com} dataset and show how the attention weights progress with respect to the occurrence time of the next event (Figure \ref{fig:case_1} and \ref{fig:case_2}). The sequence of user activities starts from the top to bottom. Each activity consists of the type of user behavior and the product.

In Figure \ref{fig:case_1}, it is evident that right after the final event, actions such as \emph{view} and \emph{search} have high attention weights, as they reflect the most immediate interests. As for the \emph{transaction} activity, the attention on \textsl{transaction-sofa} pair gradually rise from zero as time elapsed. The attention of \textsl{view-coffee table} pair increases over time as well. The patterns captured by our model are highly reasonable in e-commerce settings: 1. customers' short-term behaviors are more relevant to what they recently searched and viewed; 2. the long-term behaviors are affected by the actual purchases, and the products that they searched/viewed but haven't yet purchased.
The attention weight dynamics reflected in Figure \ref{fig:case_2} also show similar patterns. 

\subsection{Visualization of time embeddings and time kernel functions}

\begin{figure}[tbh]
    \centering
    \includegraphics[scale=0.43]{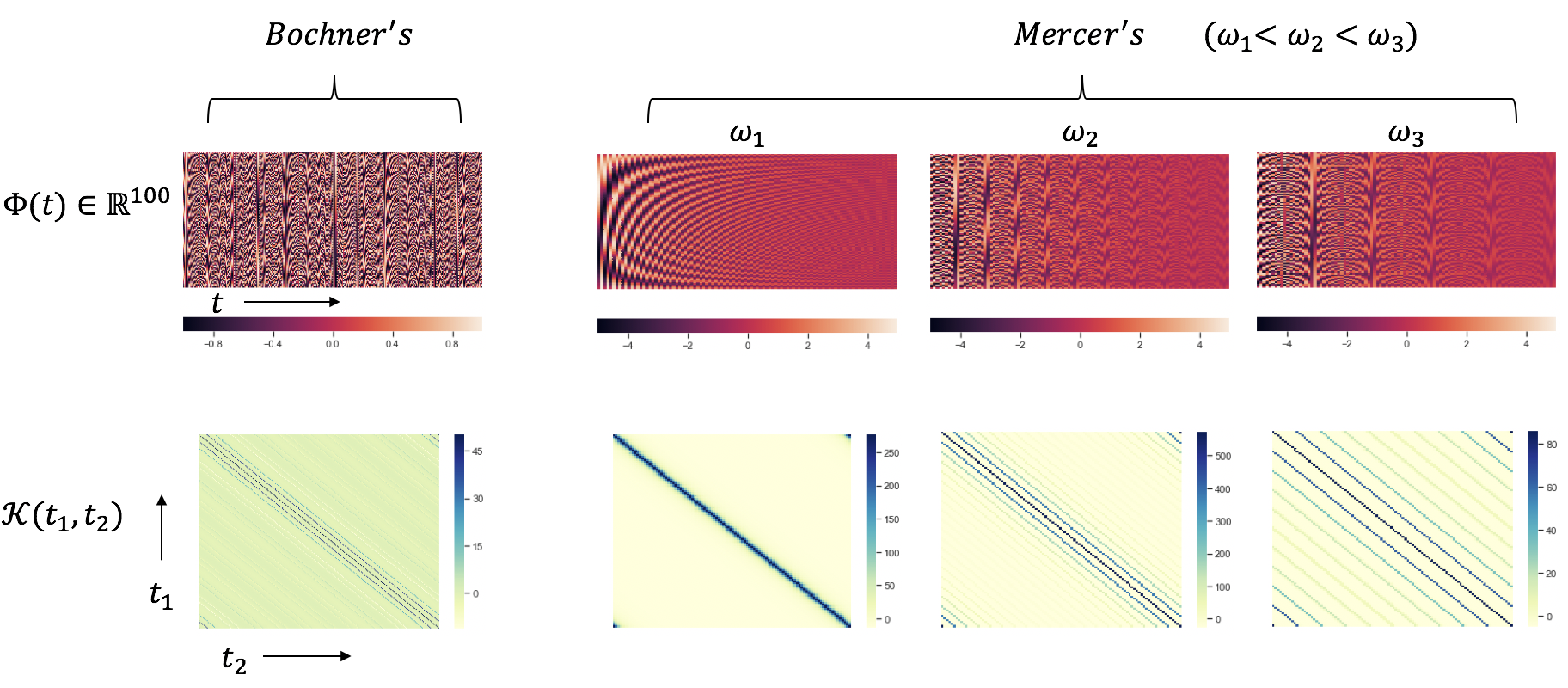}
    \caption{Visualization of the learned \textsl{Bochner non-para} and Mercer time embedding functions $\Phi$ (upper panel) and corresponding time kernel function $\mathcal{K}$ (lower panel). For the Mercer time embeddings, we sample three periodic kernels $\mathcal{K}_{\omega}$ and visualize them with their corresponding time embedding functions.}
    \label{fig:visual1}
\end{figure}

In Figure \ref{fig:visual1}, we plot the time embeddings functions $\Phi(t)$ and the corresponding kernel function $\mathcal{K}(t_1,t_2)$. Firstly, we observe that the kernel functions approximated by either Bochner time embedding or Mercer time embedding are PSD and translation-invariant (since the non-zero elements are distributed on fringes that are parallel to the diagonal in the lower panels of Figure \ref{fig:visual1}). Secondly, the visualizations show that the time embedding functions $\Phi(t)$ do capture temporal patterns, because otherwise the values in the $\Phi(t)$ matrices would be randomly distributed, as opposed to the recognizable patterns in the upper panels of Figure \ref{fig:visual1}.  

\subsection{Reference implementation}
The reference code for our implementations is provided in the supplementary material.

\end{document}